\documentclass[11pt]{article}

\usepackage[final]{acl}

\usepackage{times}
\usepackage{latexsym}

\usepackage[T1]{fontenc}

\usepackage[utf8]{inputenc}

\usepackage{microtype}

\usepackage{inconsolata}

\usepackage{graphicx}

\usepackage{booktabs}
\usepackage{xcolor}
\usepackage{multirow}
\usepackage{colortbl}
\usepackage{pifont}
\usepackage{tcolorbox}
\usepackage{listings}
\usepackage{float}
\usepackage{xspace}
\usepackage{graphicx}
\usepackage{amsmath}
\usepackage{array}
\usepackage{subcaption}
\usepackage{adjustbox}
\usepackage{makecell}
\usepackage{hhline}
\usepackage{enumitem}
\usepackage{threeparttable}
\usepackage{amssymb}
\usepackage{tikz}
\usepackage{caption}
\usepackage{tcolorbox}
\usepackage{algorithm}
\usepackage{algpseudocode}
\tcbuselibrary{skins, breakable}
\usepackage{enumitem}

\newcommand{\gpticon}{\raisebox{-0.1em}{\includegraphics[height=0.8em]{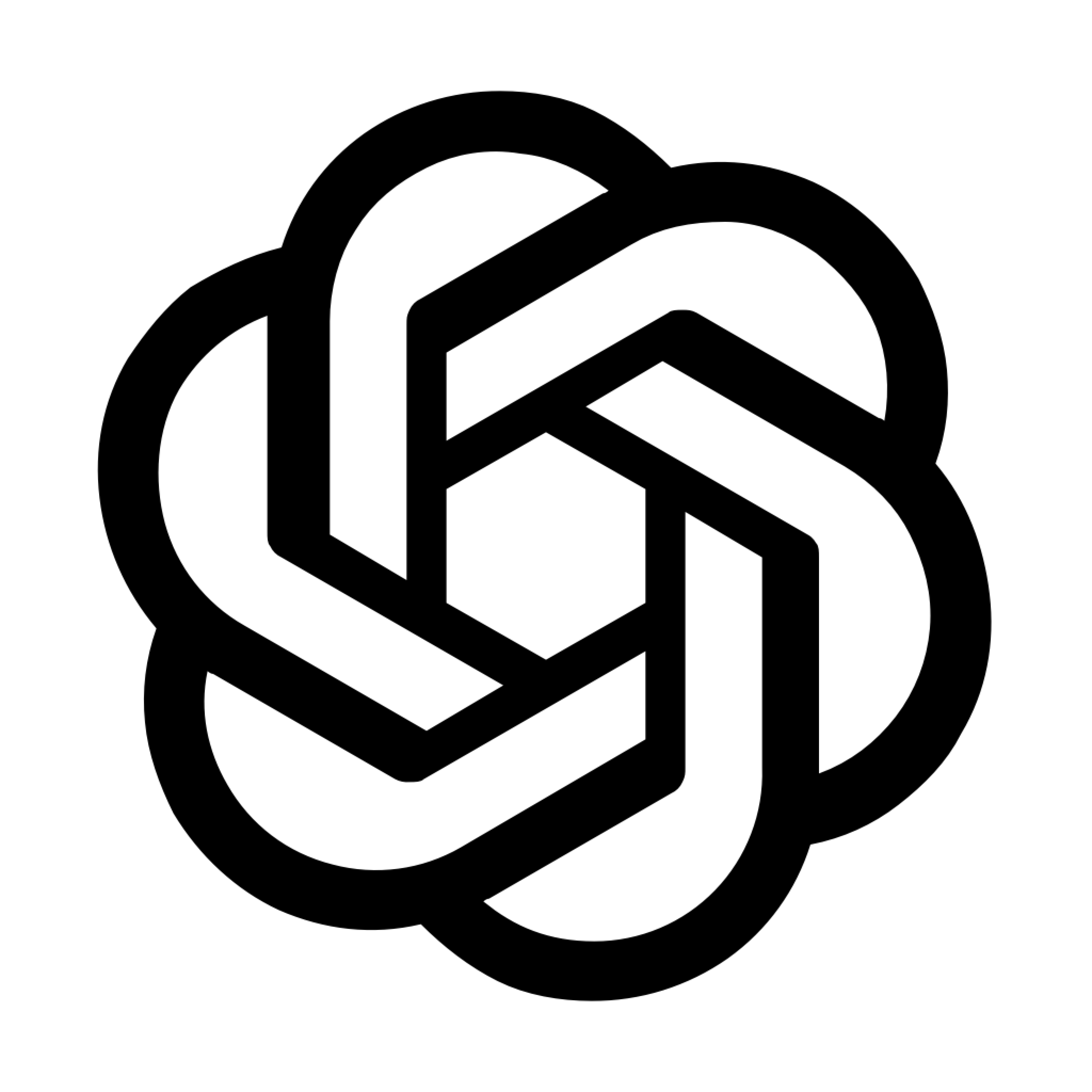}}\hspace{0.25em}}
\newcommand{\qwenicon}{\raisebox{-0.1em}{\includegraphics[height=0.8em]{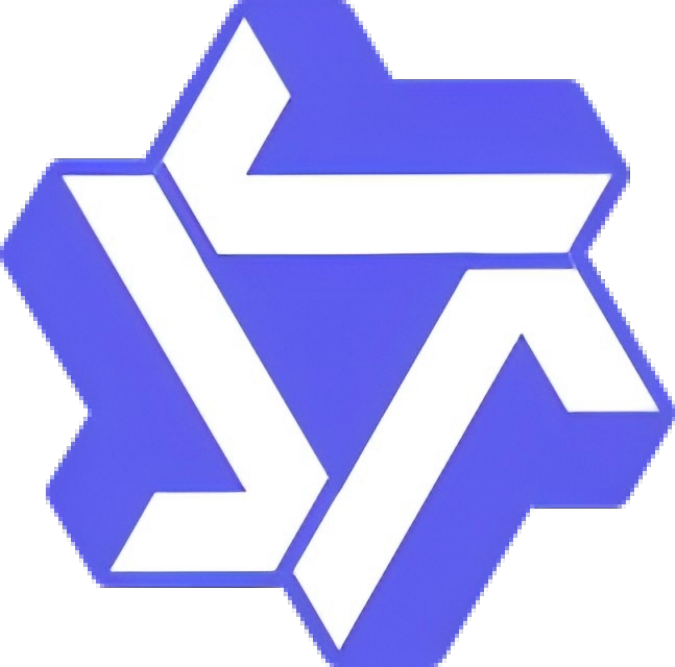}}\hspace{0.25em}}
 
\definecolor{oursbg}{RGB}{255,244,230}
\definecolor{oursbar}{RGB}{255,140,0}

\newcommand{\oursmark}{
  \makebox[3pt][l]{
    \textcolor{oursbar}{\rule[-0.4ex]{2.2pt}{2.2ex}}
    \hspace{4pt}
  }
}

\newcommand{\tabmark}{
  \makebox[3pt][l]{
    \textcolor{oursbar}{\rule[-0.4ex]{2.2pt}{2.2ex}}
    \hspace{4pt}
  }
}
\definecolor{mygreen}{HTML}{44a05c}
\definecolor{myred}{RGB}{180,53,60}
\newcommand{\cmark}{\textcolor{mygreen}{\ding{51}}}
\newcommand{\xmark}{\textcolor{myred}{\ding{55}}}

\newcommand{\model}{MemLoc\xspace}

\newcommand{\module}{Locator\xspace}

\title{Where to Look and What to Use: Retrieve–Localize–Generate for Long-Term Conversational Memory Question Answering}

\author{
\textbf{Yifan Wang\textsuperscript{1,4}\thanks{These authors contributed equally to this work.}},
\textbf{Xinkui Lin\textsuperscript{2,3}\footnotemark[1]},
\textbf{Yongxiu Xu\textsuperscript{2,3}\thanks{Corresponding authors: Yongxiu Xu, Shen Gao and Shuo Shang.}},
\textbf{Shen Gao\textsuperscript{1,4}\footnotemark[2]},
\textbf{Ruochen Yang\textsuperscript{2,3}}
\\
\textbf{Kun Huang\textsuperscript{5}},
\textbf{Yubin Wang\textsuperscript{2,3}},
\textbf{Jie Wu\textsuperscript{5}},
\textbf{Wei Liu\textsuperscript{5}},
\textbf{Jian Luan\textsuperscript{5}},
\textbf{Hongbo Xu\textsuperscript{2,3}},
\textbf{Shuo Shang\textsuperscript{1}\footnotemark[2]}
\\
\textsuperscript{1}University of Electronic Science and Technology of China, Chengdu, China \\
\textsuperscript{2}Institute of Information Engineering, Chinese Academy of Sciences, Beijing, China \\
\textsuperscript{3}School of Cyber Security, University of Chinese Academy of Sciences, Beijing, China \\
\textsuperscript{4}State Key Laboratory of Internet Architecture, Tsinghua University, Beijing, China \\
\textsuperscript{5}Independent Researcher \\
\texttt{\{yifanwang993w,jedi.shang\}@gmail.com, \{linxinkui,xuyongxiu\}@iie.ac.cn}
}

\begin{document}
\maketitle

\begin{abstract}
Retrieval-augmented generation (RAG) enables large language models (LLMs) to answer questions by accessing external knowledge and has been widely adopted for long-term conversational memory question answering. 
However, existing methods suffer from two key challenges: (1) fragmented evidence scattered across temporally distant sessions, and (2) noisy content within retrieved sessions that triggers the \textit{lost-in-the-middle} effect.
To address these challenges, we propose \model, a unified \textbf{Retrieve--Localize--Generate} framework for long-term conversational memory QA.
For retrieval, \model decomposes each session into multi-granularity memory units and performs query routing via an inner-memory graph with entropy-based granularity selection. It further models cross-session semantic and temporal dependencies through a cross-memory graph, enabling coarse-to-fine retrieval of top-$K$ relevant memory candidates.
For localization, we introduce a reasoning-based evidence locator trained with Self-reflective Hint Policy Optimization (SHPO), which performs progressive refinement by extracting query-relevant fragments within memory units to suppress noise and reranking across candidates to remove redundancy, producing a compact evidence set with lightweight location IDs.
For generation, these IDs act as precise grounding signals that guide the LLM to the correct memory positions, mitigating the lost-in-the-middle effect while preserving original contextual integrity.
Extensive experiments on four benchmarks demonstrate that \model achieves state-of-the-art retrieval accuracy and response quality while maintaining efficiency.
Our code is available at: \url{https://github.com/Nikol-coder/MemLoc}.
\end{abstract}

\section{Introduction}

Recent advances in large language models (LLMs)~\cite{li2026survey,wang2024survey,wang2025scmenhancinglargelanguage} have enabled personalized conversational assistants for applications such as customer service and intelligent agents~\cite{zhang-etal-2024-llm-based,du2025rethinkingmemoryllmbased}.
However, the limited context window of LLMs cannot accommodate continuously growing dialogue histories~\cite{hsieh2024rulerwhatsrealcontext,liu-etal-2024-lost}, making it difficult to maintain response consistency and personalization over time.
To address this, retrieval-augmented memory systems have been widely adopted, where historical dialogues are stored as external memory and selectively retrieved to ground LLM responses~\cite{qian2025memoragboostinglongcontext,yu2025memagentreshapinglongcontextllm}.
Despite their promise, existing methods face two fundamental challenges when applied to long-term conversational question answering.

\begin{figure}[ht]
  \centering
  \includegraphics[width=\linewidth]{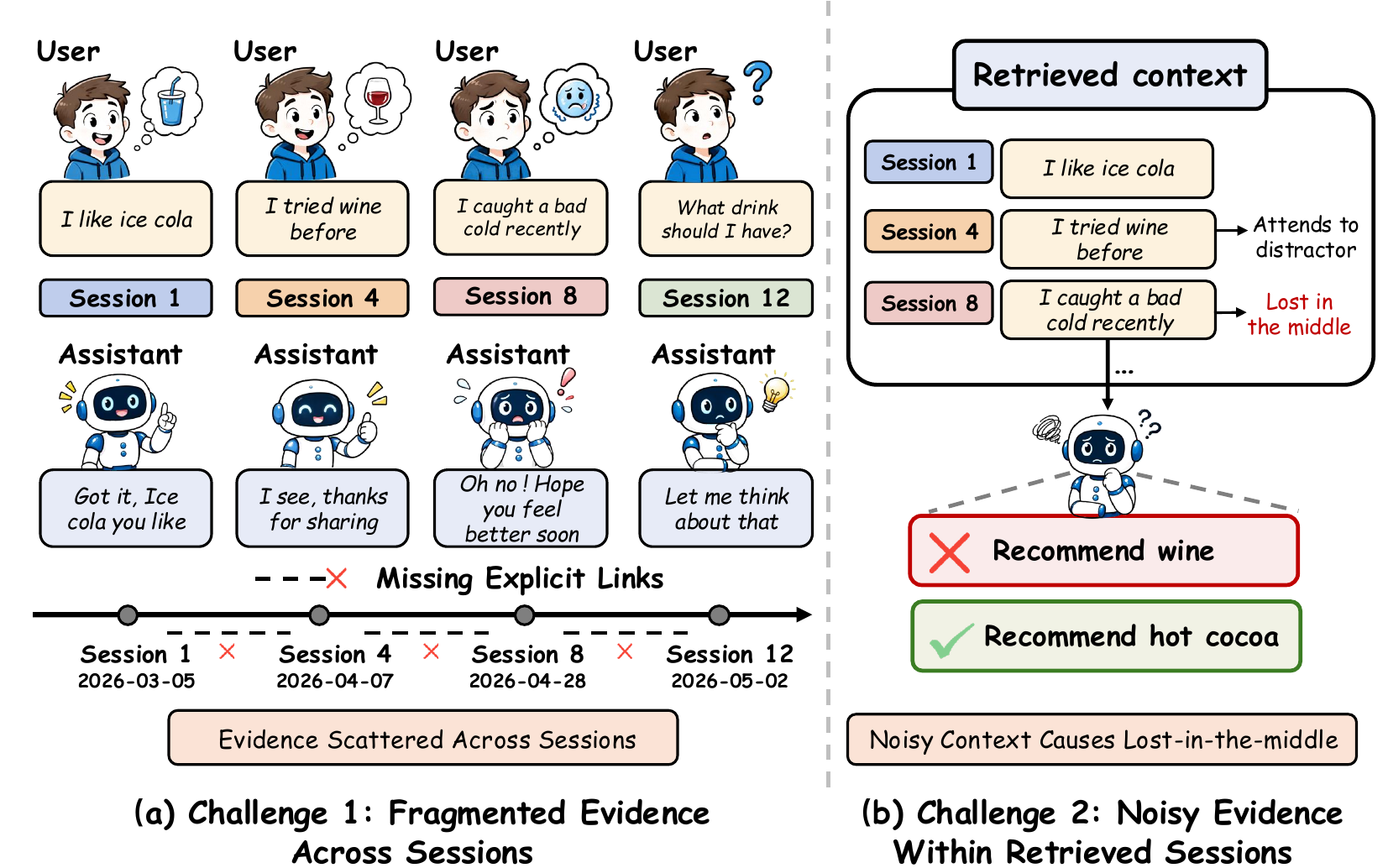}
  \caption{Challenges of Long-Term Memory Retrieval and Generation.}
  \label{fig:dongji}
\end{figure}

\textbf{Fragmented Evidence Across Sessions.}
Relevant information is often scattered across multiple temporally distant sessions, yet most retrieval methods treat each session as an independent unit.
Single-granularity approaches~\cite{lu2023memochattuningllmsuse} fail to capture fine-grained cross-session associations, while multi-granularity methods~\cite{rezazadeh2025isolatedconversationshierarchicalschemas,xu2025singlemultigranularitylongtermmemory} still lack explicit structures to model the semantic and temporal relations between related events across sessions, leading to incomplete evidence aggregation for complex multi-hop queries.

\textbf{Noisy Evidence Within Retrieved Sessions.}
Even when relevant sessions are successfully retrieved, they typically contain substantial irrelevant content alongside critical evidence.
This triggers the \textit{``lost-in-the-middle''} effect~\cite{liu-etal-2024-lost}, where LLMs attend to partial or incorrect segments.
Prior work attempts to mitigate this through filtering or memory compression~\cite{xu2025singlemultigranularitylongtermmemory,chen2025longpolongcontextselfevolution}, but such operations inevitably introduce information loss or semantic drift, compromising the fidelity of the original dialogues.

To address these challenges, we propose \model, a unified \emph{Retrieve--Localize--Generate} framework for long-term conversational memory QA.
\textbf{For retrieval}, \model organizes each session into multi-granularity memory units and builds an inner-memory graph to adaptively route queries to the most discriminative granularity via entropy-based selection.
It further constructs a cross-memory graph that captures semantic and temporal dependencies across sessions, enabling coarse-to-fine retrieval of top-$K$ candidate memories.
\textbf{For localization}, we introduce a reasoning-based evidence locator trained via Self-reflective Hint Policy Optimization (SHPO).
The locator performs two-stage evidence localization: first extracting query-relevant segments within each memory unit (inner-memory extraction) to suppress noise, then reranking across candidate units (cross-memory reranking) to eliminate redundancy, yielding a compact refined memory subset and the corresponding evidence IDs.
\textbf{For generation}, the evidence IDs serve as lightweight cues that anchor the generator to precise memory locations, mitigating the \textit{lost-in-the-middle} effect while preserving the integrity of the original memory context.

Our contributions are summarized as follows:
\begin{itemize}[leftmargin=*]
    \item We propose \model, a unified \textit{Retrieve--Localize--Generate} framework for long-term conversational memory QA. 
    It combines inner-memory routing with cross-memory propagation for coarse-to-fine retrieval, inner-memory evidence extraction with cross-memory reranking for noise-resistant evidence selection, and ID-based cues for faithful generation.
    \item We design a reasoning-based evidence locator trained via Self-reflective Hint Policy Optimization (SHPO), which contrasts the model's own correct and incorrect reasoning trajectories to self-distill strategy hints without answer leakage, enabling progressive and noise-resistant evidence localization.
    \item Extensive experiments on four long-term memory QA benchmarks demonstrate that \model achieves state-of-the-art retrieval accuracy and response quality while maintaining computational efficiency across different generator backbones.
\end{itemize}

\section{Related Work}

\paragraph{Long-Term Memory Retrieval and Generation.}
To enable personalized conversational agents, large language models (LLMs) retrieve relevant historical dialogues as external memory to ground response generation. Existing approaches mainly differ in how memory is represented and organized. Single-granularity methods treat each session as an atomic retrieval unit~\cite{zhong2024memorybank}, while multi-granularity approaches construct hierarchical structures to better capture contextual dependencies within sessions~\cite{xu2025singlemultigranularitylongtermmemory,lu2023memochattuningllmsuse}.
%
Despite these advances, long-term conversational memory QA faces two fundamental limitations.
First, relevant evidence is often distributed across temporally distant sessions, yet existing methods typically lack explicit mechanisms to model cross-session relationships, leading to incomplete evidence aggregation for complex queries. Second, retrieved memories often contain substantial redundancy and irrelevant content, which can interfere with downstream reasoning and trigger the \textit{lost-in-the-middle} effect. Prior efforts on memory compression or refinement~\cite{chen2025longpolongcontextselfevolution} attempt to mitigate this issue, but they often risk discarding fine-grained details or introducing semantic drift.
These limitations motivate more structured memory modeling and finer-grained evidence selection for long-term conversational reasoning.

\begin{figure*}[ht]
  \centering
  \includegraphics[width=0.95\linewidth]{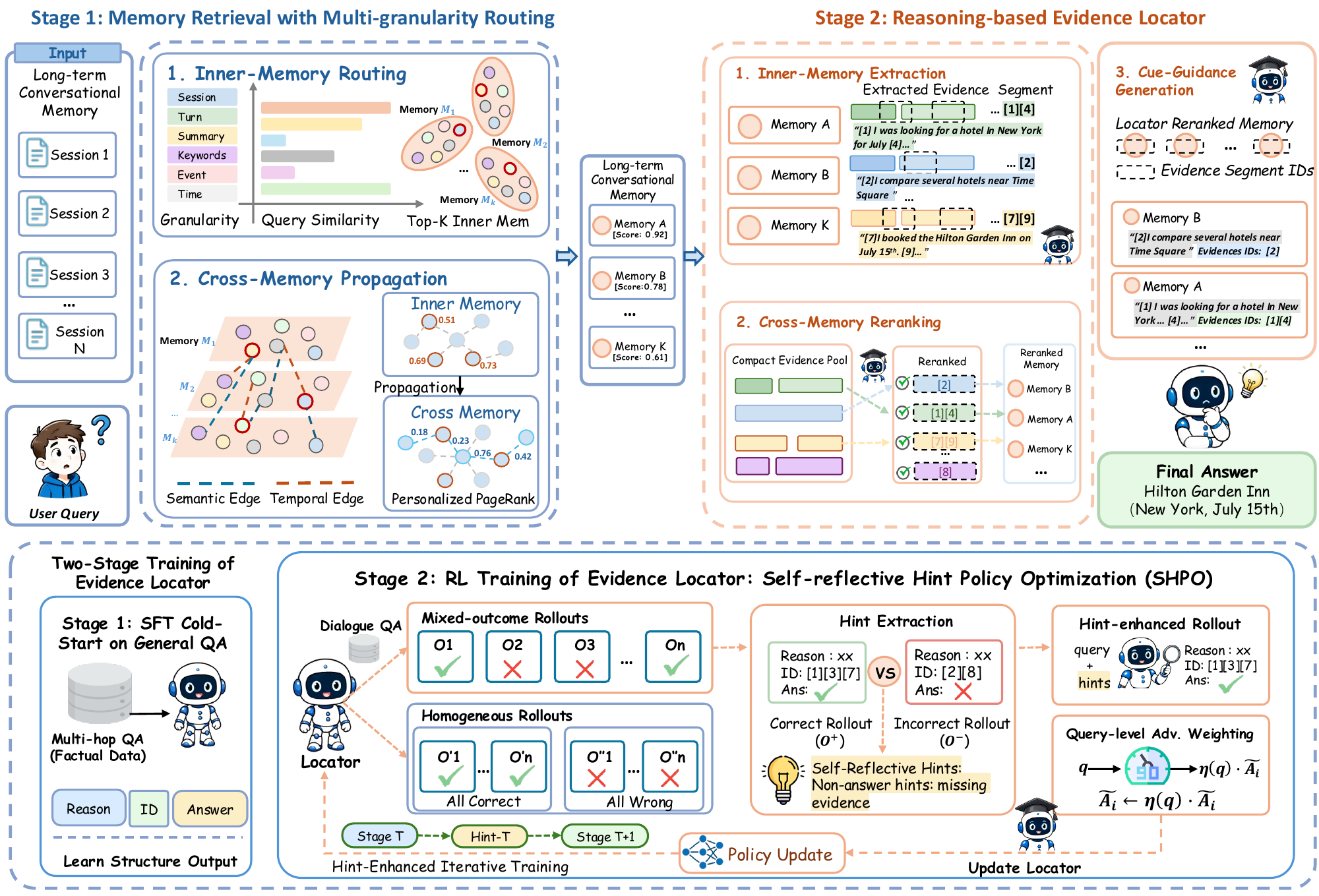}
  \caption{The overall framework of our \model.}
  \label{fig:level}
\end{figure*}

\section{Method}

\subsection{Overview}
In this section, we formalize the task of long-term conversational memory question answering 
and introduce our \model framework (Fig.~\ref{fig:level}).
Our framework consists of two core components:
(1)~\textbf{M}emory Retrieval with ~\textbf{M}ulti-granularity ~\textbf{R}outing (\textbf{MMR}, \S\ref{MMR}), 
and (2)~\textbf{R}easoning-based \textbf{E}vidence \textbf{L}ocator (\textbf{REL}, \S\ref{REL}).

\subsection{Task Formulation}

We formalize the problem as \textbf{Memory-Based Question Answering} following a \emph{Retrieve--Localize--Generate} pipeline.
Given a query $q$ and a corpus of historical memory units $\mathcal{M} = \{ M_1, M_2, \dots, M_N \}$, the objective is to (1) retrieve the most relevant memory units, (2) localize critical evidence within the retrieved units, and (3) generate a faithful answer grounded in the localized evidence.

Each memory unit $M_i$ contains a temporally coherent span of multi-turn interactions, stored as a sequence of dialogue segments:
\[
M_i = \{ u_{i,1}, u_{i,2}, \dots, u_{i,T_i} \}.
\]
The task decomposes into three stages:

\noindent (1) \textbf{Retrieve.}
A retrieval function $\mathcal{R}$ selects the top-$K$ memory units most relevant to the query $q$ from the full corpus $\mathcal{M}$:
    \begin{equation}
    \mathcal{M}_K(q) = \mathcal{R}(q, \mathcal{M}), \quad |\mathcal{M}_K(q)| = K.
    \end{equation}

\noindent (2) \textbf{Localize.}
An evidence locator $\mathcal{L}$ pinpoints the critical evidence segments within the retrieved units. It first suppresses irrelevant content within each unit (inner-memory extraction), then eliminates redundancy across units (cross-memory reranking), yielding a refined subset $\mathcal{M}_{K'}(q) \subseteq \mathcal{M}_K(q)$ with $K' \leq K$ and a set of evidence IDs that index the supporting segments:
    \begin{equation}
    \big(\mathcal{M}_{K'}(q),\; \mathcal{I}\big) = \mathcal{L}\big(q,\; \mathcal{M}_K(q)\big),
    \end{equation}
where $\mathcal{I} = \{eids_i \mid M_i \in \mathcal{M}_{K'}(q)\}$ collects the local evidence IDs for each retained memory unit.

\noindent (3) \textbf{Generate.}
A generative model $f_{\theta}$ produces the final answer $a$ by reasoning over the query $q$, the refined memory units $\mathcal{M}_{K'}(q)$, and the evidence IDs $\mathcal{I}$:
    \begin{equation}
    a = f_{\theta}\big(q,\; \mathcal{M}_{K'}(q),\; \mathcal{I}\big).
    \end{equation}
The evidence IDs serve as lightweight, interpretable cues that anchor the generator to precise memory locations. 

\subsection{Memory Retrieval with Multi-granularity Routing}
\label{MMR}

We cast conversational memory retrieval as a two-stage process:
(1) \emph{Inner-Memory Routing}, which identifies the most query-relevant fine-grained memory unit within each session; and (2) \emph{Cross-Memory Propagation}, which propagates relevance signals along semantic and temporal connections across sessions, yielding a coarse top-$K$ session candidate set.

\subsubsection{Multi-Granularity Memory Representation}
\label{sec:memory}

Each session $s_i$ is decomposed into six complementary levels of granularity: 
%
\begin{equation}
s_i =
\left(
\substack{
m_i^{\text{session}},\; m_i^{\text{turn}},\; m_i^{\text{summary}},\;
m_i^{\text{keyword}},\; m_i^{\text{event}},\; m_i^{\text{time}}
}
\right),
\end{equation}
representing session-level, turn-level, summary, keyword, event, and temporal information.
Each granularity is encoded into a dense vector $\mathbf{e}_i^g \in \mathbb{R}^d$ using the embedding model, providing a unified multi-view representation of the session.

\subsubsection{Stage 1: Inner-Memory Routing}
\label{sec:inner}

Queries may target factual details, temporal cues, or broad topics, each naturally aligning with different memory granularities.
To adaptively select the most informative granularity per query, we introduce an entropy-based routing mechanism~\cite{thiombiano2025moxemixturexlstmexperts}.

Given a query embedding $\mathbf{q}$, we compute the similarity between $\mathbf{q}$ and each memory unit of granularity $g$:
\begin{equation}
s_i^g = \mathbf{q} \cdot (\mathbf{e}_i^g)^\top .
\end{equation}
We apply a temperature-scaled softmax over all sessions to obtain a relevance distribution 
$p^g = \operatorname{softmax}_{i}(s^g / \tau)$
The sharpness of this distribution is quantified by Shannon entropy:
\begin{equation}
H^g = -\sum\nolimits_{j} p_j^g \log p_j^g .
\label{eq:entropy}
\end{equation}
A low $H^g$ indicates that only a few sessions receive high probability, i.e., granularity $g$ provides a peaked and discriminative signal.
We normalize the entropy by the logarithm of the number of sessions to account for different candidate-set sizes, and define the routing weight as
\begin{equation}
w^g =
\operatorname{softmax}_{g}
\left(
1 - \frac{H^g}{\log N_g}
\right),
\label{eq:weight}
\end{equation}
where $N_g$ denotes the number of memory units at granularity $g$.
This formulation assigns larger weights to granularities with lower normalized entropy while guaranteeing non-negative weights.
The inner-memory relevance of session $i$ is then computed as the weighted sum of its unit-level similarities:
\begin{equation}
r_i = \sum_{g=1}^{6} w^g \cdot s_i^g .
\label{eq:inner}
\end{equation}
This can be interpreted as a soft selection over memory units within each session, where the resulting score $r_i$ serves as a seed signal for cross-session propagation in the next stage.

\subsubsection{Stage 2: Cross-Memory Propagation}
\label{sec:cross}

Conversational sessions are inherently interdependent, as they share recurring entities, semantic themes, and temporal continuity. 
This implies that a memory unit highly relevant to a query often serves as a strong indicator of related evidence distributed across other sessions connected via semantic or temporal relations.
To capture these dependencies, we construct a memory graph over memory units and propagate relevance signals using Personalized PageRank~\cite{Yang_2024}, enabling global information diffusion across sessions.

\subsubsection{Memory Graph Construction}

We build an undirected graph $\mathcal{G} = (\mathcal{V}, \mathcal{E})$ where each node is a single memory unit $m_i^g$.
Edges encode two types of cross-session relations:

\paragraph{Semantic edges.}
We link memory units of the same granularity that exhibit high semantic similarity, using an adaptive threshold based on Gaussian Mixture Models (GMM)~\cite{10.5555/3009657.3009736}.
For each node, we compute its cosine similarity against all existing same-granularity nodes, take the top-$K_{\text{cand}}$ candidates, and fit a two-component GMM to their similarity scores.
Only neighbors whose similarity belongs to the higher-mean component are retained.

\paragraph{Temporal edges.}
To capture chronological order, we connect the corresponding granularity units of consecutively indexed sessions:
\begin{equation}
\mathcal{E}_{\text{temp}} = \{\,(m_i^{(g)},\, m_{i+1}^{(g)}) \mid \forall\, g \in \{1,\dots,6\},\; \forall\, i\,\},
\label{eq:temp_edges}
\end{equation}
allowing information to flow along the timeline.

\subsubsection{Propagation via Personalized PageRank}

We construct an initial relevance vector $\mathbf{r}$ by taking the unit-level scores $r_i^g = w^g \cdot s_i^g$ (which compose $r_i$ in Eq.~\ref{eq:inner}) and retaining only the top-$K_0$ entries, setting the rest to zero.
This sparsification preserves the strongest query-responsive units and reduces computation.
We then propagate these signals over $\mathcal{G}$ using Personalized PageRank (PPR):
\begin{equation}
\boldsymbol{\pi} = (1 - d) \cdot \mathbf{A} \, \boldsymbol{\pi} + d \cdot \mathbf{r},
\label{eq:ppr}
\end{equation}
where $\mathbf{A}$ is the column-stochastic transition matrix defined as
\begin{equation}
\mathbf{A}_{ij} = \begin{cases}
\frac{1}{|\mathcal{N}(j)|}, & \text{if } (i,j) \in \mathcal{E}, \\
0, & \text{otherwise},
\end{cases}
\end{equation}
$\mathcal{N}(j)$ is the neighbor set of node $j$, and $d$ is the restart probability.
The stationary distribution $\boldsymbol{\pi}$ assigns a final relevance score $\pi_i^g$ to each memory unit, effectively diffusing the query's attention along both semantic associations and temporal flows.
Finally, session-level scores are obtained by aggregating node scores:
\begin{equation}
S_i = \sum_{g=1}^{6} \pi_i^g .
\end{equation}
The top-$K$ sessions ranked by $S_i$ form the coarse recall set, which is subsequently passed to the fine-grained localization and answer generation module.

\subsection{Reasoning-based Evidence \module}
\label{REL}

Recent methods~\cite{li2025memosmemoryosai,nan2025nemoriselforganizingagentmemory} employ LLM-based filters to reduce redundancy by selecting query-relevant segments from individual memory units. However, in long-term conversational memory QA, critical evidence is often distributed across temporally dispersed dialogue sessions and requires aggregation to form complete answers. Existing filters, primarily designed for single-session contexts, lack mechanisms to identify and synthesize cross-session evidence, frequently missing crucial contextual links.

To address this limitation, we propose a reasoning-based evidence \module that operates over retrieved memory units via a two-level design: Inner-Memory Extraction and Cross-Memory Reranking. At the inner-memory level, the \module is applied to the candidate memory units retrieved by MMR to remove irrelevant content within each unit. At the cross-memory level, the \module leverages the compact evidence to rerank candidate memory units. This evidence-guided reranking yields a final subset of memory units, which is used as input to the generator.

\textbf{Inner-Memory Extraction.}
To yield a structured inner-memory representation, each retrieved memory unit $M_i \in \mathcal{M}_K(q)$ is decomposed into fine-grained atomic evidence segments, each with a unique local evidence ID:
$$\mathbf{M}_i = \{m_{i,1}, m_{i,2}, \dots, m_{i,L_i}\}$$
where $m_{i,l}$ denotes the $l$-th atomic evidence segment extracted from memory $M_i$, and $L_i$ is the total number of atomic segments within memory unit $M_i$.

For each retrieved memory unit $\mathbf{m}_i$, the \module identifies relevant evidence by outputting a set of \emph{local evidence IDs}:
\begin{equation}
eid_{s_i} = \mathrm{\module}(q, \mathbf{M}_i),
\quad eid_{s_i} \subseteq \{1,\ldots,L_i\},
\end{equation}
where each ID corresponds to atomic evidence segments directly relevant to the query. Then we construct a purified evidence block by concatenating only the selected evidence units:
\begin{equation}
B_i = \{\, m_{i,t} \mid t\in eid_{s_i} \,\},
\end{equation}
This inner-memory evidence extraction operates on all $K$ retrieved memories, effectively eliminating irrelevant content within each memory unit.

\textbf{Cross-Memory Reranking.}
We first renumber and integrate all evidence blocks $B_i$ into a compact cross-memory representation:
\begin{equation}
M^{\mathrm{in}} = \{B_1, B_2, \dots, B_K\},
\end{equation}
The \module then processes $M^{\mathrm{in}}$ to identify the minimal evidence chain required for answering $q$:
\begin{equation}
sid_s = \mathrm{\module}(q, M^{\mathrm{in}}),
\quad sid_s \subseteq \{1,\ldots,K\},
\end{equation}
\begin{equation}
\mathcal{M}_{K'}(q) = \{\, M_i \mid i \in sid_s \,\},
\end{equation}
which retains the essential evidence chains while eliminating redundant information.
Finally, we use $\mathcal{M}_{K'}(q)$ to rerank the original top-$K$ candidate memory units, producing reranked memory sets.

\textbf{Cue-Guidance Generation.}
Following cross-memory reranking, we provide the generator with both the selected memory units $\mathcal{M}_{K'}(q)$ and their corresponding evidence IDs:
\begin{equation}
eid_s = \{\,eid_{s_i} \mid M_i \in \mathcal{M}_{K'}(q)\,\}.
\end{equation}
The final response is then generated as:
\begin{equation}
a = \mathrm{Generator}\big(q, \mathcal{M}_{K'}(q), eid_s\big).
\end{equation}
These evidence IDs serve as lightweight, interpretable cues that directly anchor the generator to critical evidence segments within memory.

\subsection{Two-Stage Training of Evidence \module}

\subsubsection{Stage 1: SFT Cold-Start}
\label{sec:sft}

To bootstrap the \module with structured reasoning and accurate evidence localization before RL, we first perform supervised fine-tuning (SFT) on multi-hop QA datasets, including HotpotQA~\cite{yang2018hotpotqa}, MuSiQue~\cite{trivedi2022musique}, and 2WikiMultihopQA~\cite{ho2020constructing}.
We synthesize structured reasoning demonstrations: given a query $q$, candidate evidence set $\mathcal{M} = \{m_1, \dots, m_k\}$, and gold answer $a$, the oracle produces a target tuple $y = \langle r, \mathcal{I}, a \rangle$, where $r$ is a chain-of-thought trace explicitly referencing evidence content, $\mathcal{I} \subseteq \{1, \dots, k\}$ is the set of supporting evidence indices, and $a$ is the answer.
We then fine-tune the \module on $\mathcal{D}_{\text{sft}} = \{(x, y)\}$ with $x = (q, \mathcal{M})$, minimizing:
\begin{equation}
\mathcal{L}_{\text{sft}}(\theta) = -\mathbb{E}_{(x,y) \sim \mathcal{D}_{\text{sft}}} \left[ \log p_\theta(y \mid x) \right].
\label{eq:sft_loss}
\end{equation}
This stage instills structured reasoning and evidence localization capabilities, providing a stable starting point for the RL phase.

\subsubsection{Stage 2: Self-reflective Hint Policy Optimization}
\label{sec:shpo}

While SFT equips the \module with basic reasoning and evidence localization capabilities, long-term dialogue QA poses additional challenges that require targeted adaptation: (1) redundant evidence, (2) conflicting memory units, and (3) multi-turn dependencies.
To address these, we propose \textbf{Self-reflective Hint Policy Optimization (SHPO)}, a novel reinforcement learning strategy that lets the model act as its own teacher by contrasting its successful and failed trajectories on the same sample, and then injects the resulting hints into the next training iteration.
We conduct SHPO on out-of-domain dialogue datasets such as Time-Dialogue~\cite{du2025memoryt1reinforcementlearningtemporal,qin-etal-2021-timedial} and HaluMem~\cite{chen2026halumemevaluatinghallucinationsmemory}, which provide diverse, long-term conversations for learning robust hint generation and evidence refinement.

\paragraph{Reward Design.}
We combine three complementary reward signals to jointly optimize structured reasoning, evidence localization, and answer correctness.

\begin{itemize}[left=0pt, noitemsep, topsep=0pt, partopsep=0pt]
    \item \textbf{Format Reward} $r_{\text{fmt}}$: equals $1$ if the output contains correctly delimited tags (\texttt{<reason>}, \texttt{<id>}, \texttt{<answer>}), and $0$ otherwise.
    \item \textbf{Evidence Reward} $r_{\text{evid}}$: measures the accuracy of evidence localization by comparing the predicted evidence index set $\mathcal{I}_{\text{pred}}$ against the gold set $\mathcal{I}_{\text{gold}}$:
    \begin{equation}
        \label{eq:r_evid}
        r_{\text{evid}}
        =
        \frac{
        |\mathcal{I}_{\text{pred}}
        \cap
        \mathcal{I}_{\text{gold}}|
        }{
        |\mathcal{I}_{\text{gold}}|
        }.
    \end{equation}
    \item \textbf{Answer Reward} $r_{\text{ans}}$: equals $1$ if the gold answer $a_{\text{gold}}$ is a substring of the predicted answer $a_{\text{pred}}$, and $0$ otherwise.
    \begin{equation}
        r_{\text{ans}}
        =
        \mathbb{I}
        \big(
        a_{\text{gold}}
        \subseteq
        a_{\text{pred}}
        \big),
    \end{equation}
\end{itemize}
The final reward is computed as a weighted combination:
\begin{equation}
r
=
\alpha \cdot r_{\text{ans}}
+
\beta \cdot r_{\text{evid}}
+
(1-\alpha-\beta) \cdot r_{\text{fmt}},
\end{equation}
where $\alpha$ and $\beta$ balance answer correctness and evidence localization quality.

\paragraph{Mixed Rollout and Hint Extraction.}
Given a query $q$ and memory set $\mathcal{M}$, we sample $G$ rollouts $\{o_i\}_{i=1}^G$ from the current policy $\pi_\theta$; each outputs a structured tuple $\langle r, \mathcal{I}, a \rangle$.
A rollout is \emph{correct} if $r_{\text{evid}} = 1$ (Eq.~\ref{eq:r_evid}), and \emph{incorrect} otherwise.
We focus on \textbf{mixed-outcome} samples: those where at least one rollout is correct and at least one is incorrect.
For each such instance, the policy itself acts as a teacher $\mathcal{T}$: it contrasts correct trajectories $o^{+}$ with incorrect ones $o^{-}$ and distills a concise \emph{non-answer hint} $h$ that identifies missing or misused evidence without revealing the gold answer:
\begin{equation}
h = \mathcal{T}(o^{+},\, o^{-}).
\end{equation}
Since both trajectories come from the same model, the hint captures decision-level gaps that are immediately actionable for the current policy.

\paragraph{Hint-Enhanced Iterative Training.}
Let $x^{(0)}$ denote the original prompt specifying the structured generation task.
At training stage $t$, for each mixed-outcome query we construct an augmented prompt by appending the distilled hint $h^{(t)}$:
\begin{equation}
x^{(t+1)} = x^{(t)} \oplus h^{(t)}.
\end{equation}
Non-mixed samples retain their prompts unchanged, as they offer limited contrastive signal.
We then perform GRPO on the updated prompt set; the injected hints steer the policy toward mislocalized evidence patterns exposed by rollout discrepancies.
Iterating this procedure yields a self-improving curriculum where the policy progressively sharpens its evidence localization by learning from its own success-failure contrasts.

\begin{table*}[ht]
\centering
\footnotesize
\definecolor{colorMain}{HTML}{E6F0FF}
\definecolor{colorBlock}{HTML}{F7F7F7}

\caption{
Best results are shown in \textbf{bold}, second-best are \underline{underlined}.
\textbf{4o‑J}: GPT‑4o‑as‑Judge; \textbf{F1}: token‑level F1; \textbf{Tokens}: average tokens per query.
All methods use \gpticon GPT‑4o mini as the generator.
}
\label{tab:qa_perf_top3}

\resizebox{\textwidth}{!}{
\begin{tabular}{c|ccc|ccc|ccc|ccc}
\toprule
\multirow{2}{*}{\textbf{Model}} &
\multicolumn{3}{c|}{\cellcolor{blue!10}\textbf{LongMemEval‑S}} &
\multicolumn{3}{c|}{\cellcolor{blue!10}\textbf{LongMemEval‑M}} &
\multicolumn{3}{c|}{\cellcolor{blue!10}\textbf{LoCoMo}} &
\multicolumn{3}{c}{\cellcolor{blue!10}\textbf{Long‑MT‑Bench+}} \\
\cmidrule(lr){2-4}\cmidrule(lr){5-7}\cmidrule(lr){8-10}\cmidrule(lr){11-13}
& \textbf{4o‑J} & \textbf{F1} & \textbf{Tokens}
& \textbf{4o‑J} & \textbf{F1} & \textbf{Tokens}
& \textbf{4o‑J} & \textbf{F1} & \textbf{Tokens}
& \textbf{4o‑J} & \textbf{F1} & \textbf{Tokens} \\
\midrule

Full History &
50.60 & 11.48 & 103k &
12.20 & 5.70 & 128k &
33.43 & 12.23 & 20.1k &
67.44 & 36.07 & 19.2k \\

RAPTOR &
32.20 & 12.08 & 6,254 &
\multicolumn{3}{c}{\cellcolor{gray!5}\textcolor{gray}{\textit{Timeout}}} &
31.72 & 14.55 & 1,931 &
59.72 & 37.69 & 10.6k \\

SeCom &
56.00 & 12.95 & 2,741 &
42.80 & 11.33 & 2,821 &
44.21 & 13.79 & 1,021 &
64.58 & 36.68 & 4,714 \\

A‑Mem &
55.60 & 13.73 & 9,018 &
\multicolumn{3}{c}{\cellcolor{gray!5}\textcolor{gray}{\textit{Timeout}}} &
40.81 & 14.72 & 3,042 &
65.73 & 36.82 & 13.7k \\

HippoRAG 2 &
57.60 & 14.73 & 8,530 &
\multicolumn{3}{c}{\cellcolor{gray!5}\textcolor{gray}{\textit{Timeout}}} &
45.62 & 16.66 & 2,991 &
63.54 & 35.64 & 13.6k \\

Mem0 &
42.00 & 17.72 & 6,787 &
32.00 & 14.38 & 7,452 &
36.39 & 8.16 & 2,450 &
38.33 & 25.42 & 8,100 \\

MemGAS &
60.20 & 20.38 & 8,829 &
45.40 & \underline{16.85} & 8,852 &
41.07 & 17.66 & 2,825 &
69.44 & \underline{41.49} & 12.9k \\

MemoryAgent &
42.80 & 11.71 & 11.2k &
34.20 & 9.58 & 11.8k &
38.72 & 14.80 & 6,800 &
39.03 & 11.69 & 16.2k \\

Memory‑R1 &
45.60 & \underline{21.49} & 13.5k &
29.00 & 5.95 & 14.2k &
38.07 & 5.52 & 7,900 &
55.90 & 8.46 & 18.4k \\

RecurSum &
35.40 & 12.29 & 8,853 &
23.80 & 10.04 & 8,927 &
22.56 & 9.14 & 3,074 &
24.65 & 26.58 & 13.5k \\

MemoryTree &
41.40 & 12.02 & 7,600 &
35.80 & 11.57 & 8,200 &
35.21 & 8.95 & 5,300 &
51.39 & 30.13 & 11.5k \\

MemSkill &
59.14 & 15.61 & 6,900 &
42.20 & 14.88 & 7,300 &
47.21 & 17.37 & 4,900 &
62.33 & 32.01 & 10.7k \\

\rowcolor{orange!5}
\tabmark \textbf{\model (4B)} &
\underline{65.80} & 16.80 & 8,645 &
\underline{48.20} & 13.41 & 8,326 &
\underline{57.20} & \underline{18.13} & 2,744 &
\underline{73.96} & 37.98 & 12.5k \\

\rowcolor{orange!5}
\tabmark \textbf{\model (8B)} &
\textbf{68.40} & \textbf{21.52} & 8,941 &
\textbf{54.20} & \textbf{17.27} & 8,537 &
\textbf{57.40} & \textbf{19.99} & 2,907 &
\textbf{77.42} & \textbf{46.65} & 13.9k \\

\bottomrule
\end{tabular}
}
\end{table*}

\paragraph{Policy Optimization.}
We optimize $\pi_\theta$ via GRPO, re-weighting advantages toward mixed-outcome queries.
For a group of $G$ rollouts $\{o_i\}$ from query $q$, the standard group-normalized advantage $\hat{A}_{i,t}$ is scaled by a query-dependent factor:
\begin{equation}
\tilde{A}_{i,t} = \eta(q) \cdot \hat{A}_{i,t},
\end{equation}
with $\eta(q) = 1$ for mixed-outcome queries, and a smaller value for homogeneous ones.
A KL penalty to the SFT reference $\pi_{\mathrm{sft}}$ prevents forgetting.
The per-stage objective is
\begin{equation}
\resizebox{\linewidth}{!}{%
    $\begin{aligned}
    \mathcal{J}^{(t)}(\theta) = \mathbb{E}_{q,\mathcal{M}}\bigg[
    &\frac{1}{G}\sum_{i=1}^{G} \min\!\big(\rho_i \tilde{A}_i,\; \operatorname{clip}(\rho_i, 1-\epsilon, 1+\epsilon) \tilde{A}_i\big) \\
    &\qquad - \gamma D_{\mathrm{KL}}(\pi_\theta \| \pi_{\mathrm{sft}}) \bigg],
    \end{aligned}$%
}
\label{eq:grpo_obj}
\end{equation}
where $\rho_i = \frac{\pi_\theta(o_i \mid q,\mathcal{M})}{\pi_{\mathrm{old}}(o_i \mid q,\mathcal{M})}$ is the importance sampling ratio and $\epsilon$ is the clipping parameter.

\section{Experiment}
\subsection{Datasets and Evaluation Metrics}
We evaluate our method on four long-term memory QA benchmarks: \textit{LongMemEval-S (LME-S)}, \textit{LongMemEval-M (LME-M)}~\cite{wu2025longmemevalbenchmarkingchatassistants}, \textit{LoCoMo}~\cite{maharana-etal-2024-evaluating}, and \textit{Long-MT-Bench+}~\cite{pan2025memoryconstructionretrievalpersonalized}. These benchmarks assess the long-term conversational memory capabilities of LLM agents.
We evaluate both \textbf{retrieval} and \textbf{generation} performance. 
For retrieval, we report \textbf{Recall@K} and \textbf{NDCG@K} to measure retrieval coverage and ranking quality, respectively. 
For generation, we primarily adopt LLM-as-a-Judge evaluation using \textbf{GPT-4o}~\cite{zheng2023judging} as the main evaluation protocol, assessing factual consistency, semantic alignment, and response quality with respect to reference answers. 
In addition, we report conventional automatic metrics, including \textbf{F1}, \textbf{BLEU}, and \textbf{BERTScore}, as complementary indicators~\cite{xu2025singlemultigranularitylongtermmemory,maharana-etal-2024-evaluating,pan2025memoryconstructionretrievalpersonalized}. 
More comprehensive experimental results and analyses are provided in the Table~\ref{tab:qa_perf_top3_all}.

\subsection{Baselines}

We compare \model against four categories of baselines.
\textbf{(0) Full-context reference.}
This method answers each query by processing the entire conversation history in chronological order.
\textbf{(1) Structured RAG models} construct external knowledge structures (graphs or trees) from raw conversations and retrieve over them: RAPTOR~\cite{sarthi2024raptorrecursiveabstractiveprocessing}, HippoRAG~2~\cite{gutiérrez2025hipporagneurobiologicallyinspiredlongterm}, and MemoryTree~\cite{rezazadeh2025isolatedconversationshierarchicalschemas}.
\textbf{(2) Memory-based methods} explicitly maintain a long-term memory store using summarization, semantic linking, topic segmentation, or multi-granularity indexing, and retrieve relevant entries at query time: RecurSum~\cite{wang2025recursivelysummarizingenableslongterm}, A-Mem~\cite{xu2025amemagenticmemoryllm}, SeCom~\cite{pan2025memoryconstructionretrievalpersonalized}, MemGAS~\cite{xu2025singlemultigranularitylongtermmemory}, and Mem0~\cite{chhikara2025mem0}.
\textbf{(3) RL-based memory methods} train an LLM agent with reinforcement learning to actively manage or compress memory: Memory-R1~\cite{yan2026memoryr1enhancinglargelanguage}, MemoryAgent~\cite{yu2025memagent}, and MemSkill~\cite{zhang2026memskilllearningevolvingmemory}.

\subsection{Main Results}

\textbf{Question Answering Performance.}
The long-term memory QA results are shown in Table~\ref{tab:qa_perf_top3}.
We observe that:
(1) \model achieves state-of-the-art question answering performance on the 4o-J and F1 metrics across all benchmarks, consistently outperforming existing methods while maintaining effective token consumption. Notably, both variants of our locator demonstrate strong and robust performance across datasets, with the 4B model already achieving competitive results and the 8B model further delivering state-of-the-art performance, highlighting the scalability and effectiveness of our method.
(2) Although RL-based methods exhibit strong performance on general QA tasks, memory-based methods consistently outperform RL-based memory-management methods in long-term conversational memory QA, suggesting that explicit memory mechanisms are better suited for capturing and retrieving information across extended conversational histories. 
Further analysis is provided in the Appendix.

\begin{table}[ht]
\centering
\caption{Retrieval performance. 
Full combines MMR and REL; ``w/o'' removes the specified component. 
R@k: Recall@k; N@k: NDCG@k.}
\label{tab:retrieval_performance_zhu}

\begin{adjustbox}{width=\linewidth,center}
\resizebox{\columnwidth}{!}{
\begin{tabular}{@{} p{3.25cm} c c c c c c @{}}
\toprule
\textbf{Method} & \textbf{R@3} & \textbf{N@3} & \textbf{R@5} & \textbf{N@5} & \textbf{R@10} & \textbf{N@10} \\
\midrule

\rowcolor{blue!5}
\multicolumn{7}{c}{\textbf{LongMemEval-S}} \\
\addlinespace[2pt]

SeCom        & 71.06 & 80.88 & 80.43 & 83.08 & 89.15 & 85.11 \\
MemGAS       & 78.51 & \textbf{86.83} & 88.94 & \textbf{88.77} & 94.47 & \textbf{89.96} \\

\cmidrule{1-7}

\rowcolor{orange!5}
\textbf{\oursmark \model (MMR)} 
& 83.90 & 81.02 & 90.35 & 83.38 & 94.50 & 84.02 \\
\rowcolor{orange!5}
\textbf{\oursmark \model (Full)}
& \textbf{87.97} & 86.81
& \textbf{92.56} & 88.62
& \textbf{94.50} & {89.44} \\

\rowcolor{gray!8}
\quad \textcolor{gray}{w/o IM-Routing}
& 86.02 & 83.36 & 89.98 & 85.09 & 92.60 & 87.72 \\
\rowcolor{gray!8}
\quad \textcolor{gray}{w/o CM-Prop.}
& 85.27 & 83.65 & 89.23 & 84.92 & 92.95 & 85.55 \\

\rowcolor{gray!8}
\quad \textcolor{gray}{\textit{only SFT}}
& 86.45 & 82.40 & 92.53 & 84.77 & 94.50 & 85.57 \\
\rowcolor{gray!8}
\quad \textcolor{gray}{\textit{only RL}}
& 87.25 & 86.37 & 92.08 & 88.08 & 94.50 & 89.03 \\

\midrule

\rowcolor{blue!5}
\multicolumn{7}{c}{\textbf{LoCoMo}} \\
\addlinespace[2pt]

SeCom        & 55.24 & 57.90 & 64.80 & 62.36 & 78.30 & 66.97 \\
MemGAS       & 57.30 & 58.76 & 67.32 & 63.62 & 81.82 & 68.42 \\
\cmidrule{1-7}

\rowcolor{orange!5}
\textbf{\oursmark \model (MMR)}
& {66.84} & {54.58}
& {77.10} & {58.75}
& {90.09} & {63.83} \\

\rowcolor{orange!5}
\textbf{\oursmark \model (Full)}
& \textbf{77.85} & \textbf{69.76}
& \textbf{81.72} & \textbf{71.67}
& \textbf{90.09} & \textbf{73.80} \\

\rowcolor{gray!8}
\quad \textcolor{gray}{w/o IM-Routing}
& 61.61 & 53.15 & 72.87 & 57.88 & 81.10 & 62.88 \\
\rowcolor{gray!8}
\quad \textcolor{gray}{w/o CM-Prop.}
& 63.57 & 56.00 & 75.20 & 60.81 & 85.70 & 65.18 \\
\rowcolor{gray!8}
\quad \textcolor{gray}{\textit{only SFT}}
& 77.16 & 69.66 & 81.58 & 71.40 & 90.09 & 73.58 \\
\rowcolor{gray!8}
\quad \textcolor{gray}{\textit{only RL}}
& 64.88 & 58.74 & 75.10 & 62.97 & 90.09 & 67.23 \\

\bottomrule
\end{tabular}
}
\end{adjustbox}
\end{table}

\noindent\textbf{Retrieval Performance.}
Table~\ref{tab:retrieval_performance_zhu} reports retrieval results on LongMemEval-S and LoCoMo.
\model surpasses SeCom and MemGAS, demonstrating the effectiveness of its multi‑granularity design that combines inner-memory routing with cross-memory propagation.
Removing either inner-memory routing (w/o IM-Routing) or cross-memory propagation (w/o CM-Prop.) consistently degrades recall, confirming that both components are indispensable for locating scattered evidence across sessions.
When augmented with REL as an evidence-aware reranker, \model achieves the best scores, showing that fine-grained evidence selection further refines the retrieval ranking.
Within REL, using only SFT or only RL underperforms the full two-stage training, demonstrating that both supervised cold-start and self-reflective hint optimization are critical for building a reliable locator.

\subsection{Ablation Study}

\subsubsection{Evidence Localization}
\begin{table}[ht]
\centering
\caption{Ablation study of evidence localization, with GPT-4o-as-Judge (4o-J) as the evaluation metric.}
\label{tab:locator_ablation}
\begin{adjustbox}{width=\linewidth,center}
\begin{tabular}{lccc}
\toprule
\textbf{Method} & \textbf{LME-M} & \textbf{LME-S} & \textbf{LoCoMo} \\
\midrule
\rowcolor{blue!5}
\model (Evidence only) & 36.40 & 56.60 & 41.33 \\
\model w/ MMR & 44.20 & 65.60 & 45.87 \\
\quad + Extract & 45.60 & 66.00 & 47.13 \\
\quad + Extract + Rerank & 48.40 & 66.60 & 57.00 \\
\rowcolor{orange!5}
\quad + Extract + Rerank + Cue & \textbf{51.80} & \textbf{67.20} & \textbf{57.40} \\
\bottomrule
\end{tabular}
\end{adjustbox}
\end{table}
We evaluate the contribution of each component in our evidence localization module, including \textit{Extract}, \textit{Rerank}, and \textit{Cue}.
Using only extracted evidence as model input performs poorly, indicating that discarding contextual memory leads to significant information loss.
Starting from MMR, we progressively add each component.
\textit{Extract} improves performance by filtering irrelevant content within retrieved memory units.
\textit{Rerank} further improves performance by refining cross-memory relevance estimation based on compact evidence.
Finally, \textit{Cue} provides explicit guidance to the generator, leading to the best performance across all benchmarks.
Overall, each component contributes complementary benefits, and their combination yields the most effective evidence localization and utilization.

\subsubsection{Self-reflective Hint Policy Optimization}

\begin{figure}[ht]
  \centering
  \includegraphics[width=\linewidth]{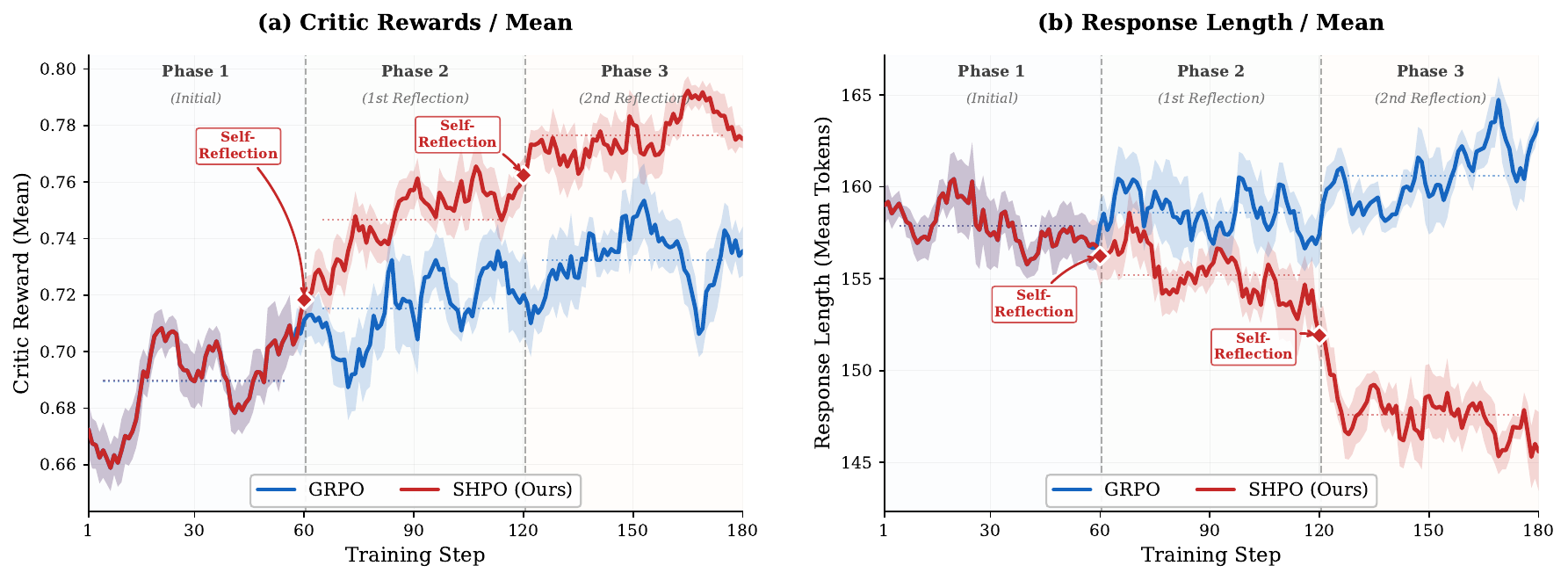}
  \caption{Training dynamics comparison between GRPO and SHPO on the 8B locator.}
  \label{fig:shpo}
\end{figure}

As shown in Fig.~\ref{fig:shpo}, SHPO introduces self-reflective hints on mixed-outcome samples at step 60, leading to a noticeable improvement in mean reward during subsequent training. 
Meanwhile, the average response length decreases significantly, indicating fewer redundant descriptions of irrelevant memory fragments and more precise evidence localization.

\begin{table}[ht]
\centering
\caption{Training ablation of the locator on LongMemEval-S.
R@k: Recall@k; N@k: NDCG@k.}
\label{tab:training_ablation}

\begin{adjustbox}{width=\linewidth,center}
\resizebox{\columnwidth}{!}{
\begin{tabular}{@{} p{3.25cm} c c c c c c @{}}
\toprule
\textbf{Method} & \textbf{R@3} & \textbf{N@3} & \textbf{R@5} & \textbf{N@5} & \textbf{R@10} & \textbf{N@10} \\
\midrule

\rowcolor{orange!5}
\textbf{\oursmark No locator}
& 83.90 & 81.02 & 90.35 & 83.38 & 94.50 & 84.02 \\

\rowcolor{gray!8}
\quad \textcolor{gray}{Base (untrained)}
& 84.62 & 81.91 & 91.68 & 84.18 & 94.50 & 84.91 \\

\rowcolor{gray!8}
\quad \textcolor{gray}{Only SFT}
& 86.45 & 82.40 & 92.53 & 84.77 & 94.50 & 85.57 \\

\rowcolor{gray!8}
\quad \textcolor{gray}{Only RL (GRPO)}
& 86.98 & 84.71 & 92.03 & 86.42 & 94.50 & 86.68 \\

\rowcolor{gray!8}
\quad \textcolor{gray}{Only RL (SHPO)}
& 87.25 & 86.37 & 92.08 & 88.08 & 94.50 & 89.03 \\

\rowcolor{gray!8}
\quad \textcolor{gray}{Full (SFT+GRPO)}
& 87.63 & 85.46 & 92.47 & 87.34 & 94.50 & 88.12 \\

\rowcolor{orange!5}
\textbf{\oursmark Full (SFT+SHPO)}
& \textbf{87.97} & \textbf{86.81}
& \textbf{92.56} & \textbf{88.62}
& 94.50 & \textbf{89.44} \\

\bottomrule
\end{tabular}
}
\end{adjustbox}
\end{table}

As shown in Table~\ref{tab:training_ablation}, even without training, the locator improves R@3 from 83.90 to 84.62 over the no-locator baseline, demonstrating that the gain primarily comes from the retrieve--localize--generate framework rather than additional QA supervision. Training further improves evidence localization, with SHPO achieving the best overall performance.

\subsection{Statistical Significance}
To assess the stability of our results, we conduct three independent runs with different random seeds.
%
%
We report the mean, standard deviation, and 95\% confidence intervals on both LongMemEval-S and LoCoMo. As shown in Table~\ref{tab:significance}, the small standard deviations demonstrate stable performance across runs, while the confidence intervals remain well above those of other methods, further confirming the robustness of our improvements.

\begin{table}[ht]
\centering
\caption{Mean $\pm$ SD and 95\% CI over three runs.}
\label{tab:significance}
\begin{adjustbox}{width=\linewidth,center}
\resizebox{\columnwidth}{!}{
\begin{tabular}{@{} l c c c @{}}
\toprule
\textbf{Setting} & \textbf{4o-J $\pm$ SD} & \textbf{95\% CI} & \textbf{F1 $\pm$ SD} \\
\midrule
LongMemEval-S
& $68.40 \pm 0.62$
& $[66.86,\,69.94]$
& $21.52 \pm 0.35$ \\
LoCoMo
& $57.40 \pm 0.42$
& $[56.36,\,58.44]$
& $19.99 \pm 0.25$ \\
\bottomrule
\end{tabular}
}
\end{adjustbox}
\end{table}

\section{Conclusion}

In this paper, we propose \model, a Retrieve--Localize--Generate framework for long-term conversational memory QA, where both retrieval and localization are built upon inner-memory and cross-memory mechanisms.
For retrieval, we first organize each session into multi-granularity memory units and construct an inner-memory graph to route queries to the most relevant memory granularity. We then model semantic and temporal relations across sessions via a cross-memory graph, enabling coarse-to-fine retrieval of top-$K$ candidate memories from long dialogue histories.
For localization, we train an evidence locator via self-reflective hint optimization. The locator first extracts query-relevant evidence within each memory unit to suppress noise and irrelevant content, and then performs cross-memory reranking to remove redundant evidence across sessions. The resulting evidence is used to guide response generation with precise memory grounding.
For generation, the generator is guided by lightweight ID-based cues that anchor it to the critical evidence within retrieved memories, yielding faithful answers.
Extensive experiments on multiple benchmarks demonstrate the effectiveness and robustness of our approach.

\section*{Limitations}
Our framework introduces additional computational overhead from cross-session graph construction, propagation, and coarse-to-fine retrieval, which may increase latency for extremely long conversations. Although top-$K$ sparsification and adaptive GMM thresholding mitigate this cost, retrieval remains slower than flat vector search. 
Given the substantial improvements in multi-hop evidence localization and noise resilience, this latency trade-off is justified. Notably, our framework is architecture-agnostic and can be seamlessly integrated with stronger base LLMs or embedders, with performance expected to scale as foundational capabilities advance. Future research will focus on incremental graph maintenance, online retrieval acceleration, and joint end-to-end optimization of the retriever and locator.


\section*{Ethics Statement}
In this paper, our experiments use only publicly available long-term memory QA benchmarks and open-source dialogue datasets. All models and APIs are used in accordance with their respective licensing and usage terms, and no private or sensitive user conversation data are involved. Therefore, we affirm that our methodology and generated outputs adhere to responsible AI practices and do not compromise user privacy or data ownership.

\section*{Acknowledgement}
This paper was supported by the National Key Research and Development Program of China (No.2025YFF0730600, No.2024YFF1401300), the National Natural Science Foundation of China (No.U25B2049, 62432002 and 62406061), and the State Key Laboratory of Internet Architecture, Tsinghua University (No. HLW2025MS10).

\bibliography{custom}

\appendix

\section{Details of the SHPO Algorithm}

\begin{algorithm}[ht]
\small
\caption{Self-reflective Hint Policy Optimization (SHPO)}
\label{alg:shpo}
\begin{algorithmic}[1]
\Require SFT policy $\pi_\theta^{(0)}$, training queries $\mathcal{Q}$, stages $T$, rollouts $G$, scaling factor $\eta_{\mathrm{homo}}$
\Ensure Locator policy $\pi_\theta^{(T)}$
\State Initialize prompts: $x_q^{(0)} \leftarrow x,\ \forall q \in \mathcal{Q}$
\For{$t = 0,\dots,T-1$}
    \For{each query $q$ with prompt $x_q^{(t)}$}
        \State Sample $G$ rollouts $\{o_i\}$ from $\pi_\theta^{(t)}(\cdot \mid x_q^{(t)})$
        \State Compute token-level rewards $r_{i,t}$ and group-normalized advantages $\hat{A}_{i,t}$
        \State Determine outcome type: mixed if $\{r_{\mathrm{evid}}\}$ contains both $0$ and $1$
        \State Set $\eta(q) \leftarrow 1$ if mixed, else $\eta_{\mathrm{homo}}$
        \State Scale advantages: $\tilde{A}_{i,t} \leftarrow \eta(q) \cdot \hat{A}_{i,t}$
    \EndFor
    \State Update policy: $\pi_\theta^{(t)} \leftarrow \arg\max_{\theta} \mathcal{J}^{(t)}(\theta)$ \Comment{GRPO step}
    \For{each mixed-outcome query $q$}
        \State Pick correct rollouts $o^{+}$ and incorrect ones $o^{-}$ from its group
        \State Extract hint: $h^{(t)} \leftarrow \mathcal{T}(o^{+}, o^{-})$
        \State Augment prompt: $x_q^{(t+1)} \leftarrow x_q^{(t)} \oplus h^{(t)}$
    \EndFor
    \State For homogeneous queries, $x_q^{(t+1)} \leftarrow x_q^{(t)}$
\EndFor
\State \Return $\pi_\theta^{(T)}$
\end{algorithmic}
\end{algorithm}

Algorithm~\ref{alg:shpo} presents the overall procedure of Self-reflective Hint Policy Optimization (SHPO). Starting from an SFT-initialized locator policy $\pi_\theta^{(0)}$, SHPO iteratively performs rollout sampling, policy optimization, and self-reflective hint augmentation. 
For each training query $q$, the current policy samples a group of $G$ reasoning rollouts conditioned on the prompt $x_q^{(t)}$. 
We then compute token-level rewards and group-normalized advantages following the GRPO framework. 

To emphasize informative training signals, SHPO distinguishes between \emph{mixed-outcome} queries and \emph{homogeneous} queries. 
A query is considered mixed-outcome if its rollout group contains both successful and failed evidence localization trajectories. 
Such cases indicate that the model already possesses partial reasoning capability but remains unstable. 
Accordingly, we preserve their original advantages, while down-weighting homogeneous queries using a scaling factor $\eta_{\mathrm{homo}}$. 
The policy is subsequently updated using the re-weighted advantages through a GRPO optimization step.

After each policy update, SHPO performs self-reflective hint extraction on mixed-outcome queries. 
Specifically, we compare correct rollouts $o^{+}$ against incorrect rollouts $o^{-}$ within the same query group and employ a teacher model $\mathcal{T}(\cdot)$ to identify the reasoning divergence that caused failure. 
The teacher model then generates a concise and domain-general corrective hint, which is appended to the original prompt for the next training stage. 
In contrast, prompts corresponding to homogeneous queries remain unchanged. 
Through this iterative process, SHPO progressively improves the model’s ability to localize evidence by leveraging self-generated reflective guidance derived from its own reasoning trajectories.

\section{Datasets}

\begin{table}[ht]
\centering
\caption{Dataset statistics. `Avg.' denotes the average per conversation.}
\resizebox{\linewidth}{!}{
\begin{tabular}{c|cccc}
\toprule
\textbf{Dataset} & \textbf{LME-S} & \textbf{LME-M} & \textbf{LoCoMo} & \textbf{Long-MT-Bench+} \\
\midrule
Total Conversations & 500 & 500 & 10 & 11 \\
Avg. Sessions       & 50.2 & 501.9 & 27.2 & 4.9 \\
Avg. Queries        & 1.0 & 1.0 & 198.6 & 26.2 \\
Avg. Tokens         & 103,137 & 1,019,117 & 20,079 & 19,195 \\
Session Dates       & \cmark & \cmark & \cmark & \xmark \\
Retrieval Ground-Truth & \cmark & \cmark & \cmark & \xmark \\
QA Ground-Truth         & \cmark & \cmark & \cmark & \cmark \\
Conversation Type       & User-AI & User-AI & User-User & User-AI \\
\bottomrule
\end{tabular}
}
\label{tab:dataset_statistics}
\end{table}

Table~\ref{tab:dataset_statistics} summarizes the statistics of all datasets, including conversation scale, session depth, and average context length.

\textbf{LongMemEval-S} and \textbf{LongMemEval-M} are multi-session User–AI conversation benchmarks designed for long-term memory reasoning and retrieval, where each query requires aggregating relevant information across multiple historical sessions with varying granularity and temporal distance.

\textbf{LoCoMo} is a User–User conversational dataset with long dialogues, where each query involves fine-grained reasoning over a large number of utterances, making it particularly challenging for retrieval and evidence localization.

\textbf{Long-MT-Bench+} is an extended version of MT-Bench~\cite{10.5555/3666122.3668142} tailored for long-context conversational evaluation, focusing on multi-turn instruction-following and reasoning under extended dialogue history.

\section{Retrieval Performance}

\begin{table}[ht]
\centering
\caption{Retrieval performance. 
Full combines MMR and REL; ``w/o'' removes the specified component. 
R@k: Recall@k; N@k: NDCG@k.}
\label{tab:retrieval_performance_2}

\begin{adjustbox}{width=\linewidth,center}
\resizebox{\columnwidth}{!}{
\begin{tabular}{@{} p{3.25cm} c c c c c c @{}}
\toprule
\textbf{Method} & \textbf{R@3} & \textbf{N@3} & \textbf{R@5} & \textbf{N@5} & \textbf{R@10} & \textbf{N@10} \\
\midrule

\rowcolor{blue!5}
\multicolumn{7}{c}{\textbf{LongMemEval-S}} \\
\addlinespace[2pt]

SeCom        & 71.06 & 80.88 & 80.43 & 83.08 & 89.15 & 85.11 \\
MemGAS       & 78.51 & \textbf{86.83} & 88.94 & \textbf{88.77} & 94.47 & \textbf{89.96} \\

\cmidrule{1-7}

\rowcolor{orange!5}
\textbf{\oursmark \model (MMR)} 
& 83.90 & 81.02 & 90.35 & 83.38 & 94.50 & 84.02 \\
\rowcolor{orange!5}
\textbf{\oursmark \model (Full)}
& \textbf{87.97} & 86.81
& \textbf{92.56} & 88.62
& \textbf{94.50} & {89.44} \\

\rowcolor{gray!8}
\quad \textcolor{gray}{w/o IM-Routing}
& 86.02 & 83.36 & 89.98 & 85.09 & 92.60 & 87.72 \\
\rowcolor{gray!8}
\quad \textcolor{gray}{w/o CM-Prop.}
& 85.27 & 83.65 & 89.23 & 84.92 & 92.95 & 85.55 \\

\rowcolor{gray!8}
\quad \textcolor{gray}{\textit{only SFT}}
& 86.45 & 82.40 & 92.53 & 84.77 & 94.50 & 85.57 \\
\rowcolor{gray!8}
\quad \textcolor{gray}{\textit{only RL}}
& 87.25 & 86.37 & 92.08 & 88.08 & 94.50 & 89.03 \\

\midrule

\rowcolor{blue!5}
\multicolumn{7}{c}{\textbf{LongMemEval-M}} \\
\addlinespace[2pt]

SeCom   & 44.26 & 56.61 & 55.32 & 60.76 & 66.60 & 63.78 \\
MemGAS  & 51.06 & 61.36 & 63.62 & \textbf{66.07} & \textbf{77.02} & \textbf{69.46} \\
\cmidrule{1-7}

\rowcolor{orange!5}
\textbf{\oursmark \model (MMR)}
& {55.92} & {60.70}
& {65.61} & 63.65
& {72.16} & 65.04 \\

\rowcolor{orange!5}
\textbf{\oursmark \model (Full)}
& \textbf{63.41} & \textbf{61.58}
& \textbf{69.01} & 63.75
& {72.16} & 65.01 \\

\rowcolor{gray!8}
\quad \textcolor{gray}{w/o IM-Routing}
& 62.53 & 60.48 & 67.91 & 63.33 & 70.35 & 64.49 \\
\rowcolor{gray!8}
\quad \textcolor{gray}{w/o CM-Prop.}
& 61.92 & 60.70 & 67.61 & 63.65 & 70.97 & 64.04 \\
\rowcolor{gray!8}
\quad \textcolor{gray}{\textit{only SFT}}
& 63.30 & 62.21 & 68.97 & 64.41 & 72.16 & 65.70 \\
\rowcolor{gray!8}
\quad \textcolor{gray}{\textit{only RL}}
& 60.18 & 58.95 & 66.55 & 61.70 & 72.16 & 63.95 \\

\midrule

\rowcolor{blue!5}
\multicolumn{7}{c}{\textbf{LoCoMo}} \\
\addlinespace[2pt]

SeCom        & 55.24 & 57.90 & 64.80 & 62.36 & 78.30 & 66.97 \\
MemGAS       & 57.30 & 58.76 & 67.32 & 63.62 & 81.82 & 68.42 \\
\cmidrule{1-7}

\rowcolor{orange!5}
\textbf{\oursmark \model (MMR)}
& {66.84} & {54.58}
& {77.10} & {58.75}
& {90.09} & {63.83} \\

\rowcolor{orange!5}
\textbf{\oursmark \model (Full)}
& \textbf{77.85} & \textbf{69.76}
& \textbf{81.72} & \textbf{71.67}
& \textbf{90.09} & \textbf{73.80} \\

\rowcolor{gray!8}
\quad \textcolor{gray}{w/o IM-Routing}
& 61.61 & 53.15 & 72.87 & 57.88 & 81.10 & 62.88 \\
\rowcolor{gray!8}
\quad \textcolor{gray}{w/o CM-Prop.}
& 63.57 & 56.00 & 75.20 & 60.81 & 85.70 & 65.18 \\
\rowcolor{gray!8}
\quad \textcolor{gray}{\textit{only SFT}}
& 77.16 & 69.66 & 81.58 & 71.40 & 90.09 & 73.58 \\
\rowcolor{gray!8}
\quad \textcolor{gray}{\textit{only RL}}
& 64.88 & 58.74 & 75.10 & 62.97 & 90.09 & 67.23 \\

\bottomrule
\end{tabular}
}
\end{adjustbox}
\end{table}

Table~\ref{tab:retrieval_performance_2} reports retrieval results on LongMemEval-S, LongMemEval-M, and LoCoMo.
Across all benchmarks, \model consistently outperforms SeCom and MemGAS, demonstrating the effectiveness and generalizability of its multi-granularity memory retrieval framework.
In particular, the full model achieves substantial gains on challenging multi-session datasets such as LongMemEval-M and LoCoMo, indicating that jointly modeling inner-memory Routing and cross-memory propagation is crucial for retrieving dispersed evidence over long conversational histories.

Removing either inner-memory routing (w/o IM-Routing) or cross-memory propagation (w/o CM-Prop.) consistently degrades retrieval performance across datasets, confirming that both modules contribute complementary capabilities.
Specifically, IM-Routing improves fine-grained evidence localization within individual memories, while CM-Propagation enhances cross-memory evidence aggregation and long-range contextual propagation.
The degradation is particularly pronounced on LoCoMo, where evidence is highly scattered across long interactions, highlighting the necessity of both mechanisms for robust long-context retrieval.

When augmented with REL as an evidence-aware reranker, \model (Full) consistently improves over the MMR-only variant, especially on R@3 and N@3, demonstrating that fine-grained evidence localization effectively refines the retrieval ranking.
Within REL, using only SFT or only RL generally underperforms the complete two-stage optimization strategy.
This further validates the importance of combining supervised cold-start training with Self-reflective Hint Policy Optimization (SHPO), where SFT provides a stable initialization, while RL-based reflective optimization further enhances evidence integration and improves ranking consistency.

\section{Ablation of Multi-granularity Memory Retrieval}

\begin{table}[ht]
\centering
\caption{Ablation study of multi-granularity memory retrieval on LongMemEval-S. (\cmark: enabled, \xmark: disabled).}
\label{tab:ablation_check}
\resizebox{\columnwidth}{!}{
\begin{tabular}{l *{6}{c} c}

\toprule
\textbf{Variant} & \textbf{Sess.} & \textbf{Turn} & \textbf{Event} & \textbf{Summ.} & \textbf{Key.} & \textbf{Time} & \textbf{R@3} \\
\midrule

\rowcolor{gray!10}
\textbf{Full Method (Ours)} & \cmark & \cmark & \cmark & \cmark & \cmark & \cmark & \textbf{87.97} \\
\midrule
\multicolumn{8}{l}{\textit{A. Single-Granularity Baselines}} \\
Session Only & \cmark & \xmark & \xmark & \xmark & \xmark & \xmark & 76.18 \\
Turn Only & \xmark & \cmark & \xmark & \xmark & \xmark & \xmark & 83.34 \\
Event Only & \xmark & \xmark & \cmark & \xmark & \xmark & \xmark & 85.15 \\
Summary Only & \xmark & \xmark & \xmark & \cmark & \xmark & \xmark & 82.94 \\
Keyword Only & \xmark & \xmark & \xmark & \xmark & \cmark & \xmark & 80.21 \\
Time Only & \xmark & \xmark & \xmark & \xmark & \xmark & \cmark & 6.44 \\
\midrule
\multicolumn{8}{l}{\textit{B. Complementary Granularity Pairs}} \\
Session + Turn & \cmark & \cmark & \xmark & \xmark & \xmark & \xmark & 80.05 \\
Session + Event & \cmark & \xmark & \cmark & \xmark & \xmark & \xmark & 85.47 \\
Session + Summary & \cmark & \xmark & \xmark & \cmark & \xmark & \xmark & 83.72 \\
Turn + Event & \xmark & \cmark & \cmark & \xmark & \xmark & \xmark & 86.00 \\
Event + Summary & \xmark & \xmark & \cmark & \cmark & \xmark & \xmark & 85.90 \\
\midrule
\multicolumn{8}{l}{\textit{C. Progressive Component Addition}} \\
Turn + Event & \xmark & \cmark & \cmark & \xmark & \xmark & \xmark & 86.00 \\
\quad + Session & \cmark & \cmark & \cmark & \xmark & \xmark & \xmark & 85.65 \\
\quad + Summary & \cmark & \cmark & \cmark & \cmark & \xmark & \xmark & 87.55 \\
\quad + Keyword & \cmark & \cmark & \cmark & \xmark & \cmark & \xmark & 86.10 \\
\midrule
\multicolumn{8}{l}{\textit{D. Ablation of Specific Components}} \\
Full w/o Session & \xmark & \cmark & \cmark & \cmark & \cmark & \cmark & 86.07 \\
Full w/o Turn & \cmark & \xmark & \cmark & \cmark & \cmark & \cmark & 86.67 \\
Full w/o Event & \cmark & \cmark & \xmark & \cmark & \cmark & \cmark & 85.37 \\
Full w/o Time & \cmark & \cmark & \cmark & \cmark & \cmark & \xmark & 87.60 \\
\bottomrule
\end{tabular}
}
\end{table}

We conduct an ablation study on LongMemEval-S to analyze the contribution of different memory granularities.
As shown in Table~\ref{tab:ablation_check}, among all single-granularity baselines (\textit{Variant A}), event-level memory (\textbf{Event Only}) achieves the best performance, underscoring its central role in long-term memory retrieval.
In contrast, session-level memory (\textbf{Session Only}) performs substantially worse than finer-grained representations, indicating that excessive noise from overly broad contexts severely degrades retrieval accuracy.
Notably, time-level memory (\textbf{Time Only}) is insufficient for accurate retrieval, revealing that temporal signals alone lack semantic grounding and become effective only when combined with other memory granularities.
However, this result does not imply that temporal modeling is ineffective. Instead, temporal cues such as ``last week'' ``last month'' and ``yesterday'' can effectively narrow the search space to a manageable set of candidate sessions, serving as useful constraints when combined with semantically grounded memory representations.

Furthermore, our examination of complementary granularity combinations (\textit{Variant B \& C}) shows that different memory granularities provide complementary retrieval signals.
In particular, combining turn-level and event-level memories yields strong performance, suggesting that fine-grained conversational details and structured event abstractions jointly improve evidence localization.

Finally, removing individual components from the full model (\textit{Variant D}) consistently degrades performance.
Notably, removing event-level memory leads to the largest drop, confirming that event-centric memory organization is critical for modeling cross-memory semantic associations and tracking event evolution across long dialogue histories.
Although temporal information contributes relatively smaller gains, it remains beneficial for resolving time-sensitive and temporally conflicting queries.
Overall, these results validate the effectiveness of our multi-granularity retrieval framework, where different memory granularities collaboratively provide complementary signals for coarse-to-fine evidence localization and cross-memory reasoning.

\section{Comparison of Different Query Types}
\begin{figure}[ht]
  \centering
  \includegraphics[width=0.7\linewidth]{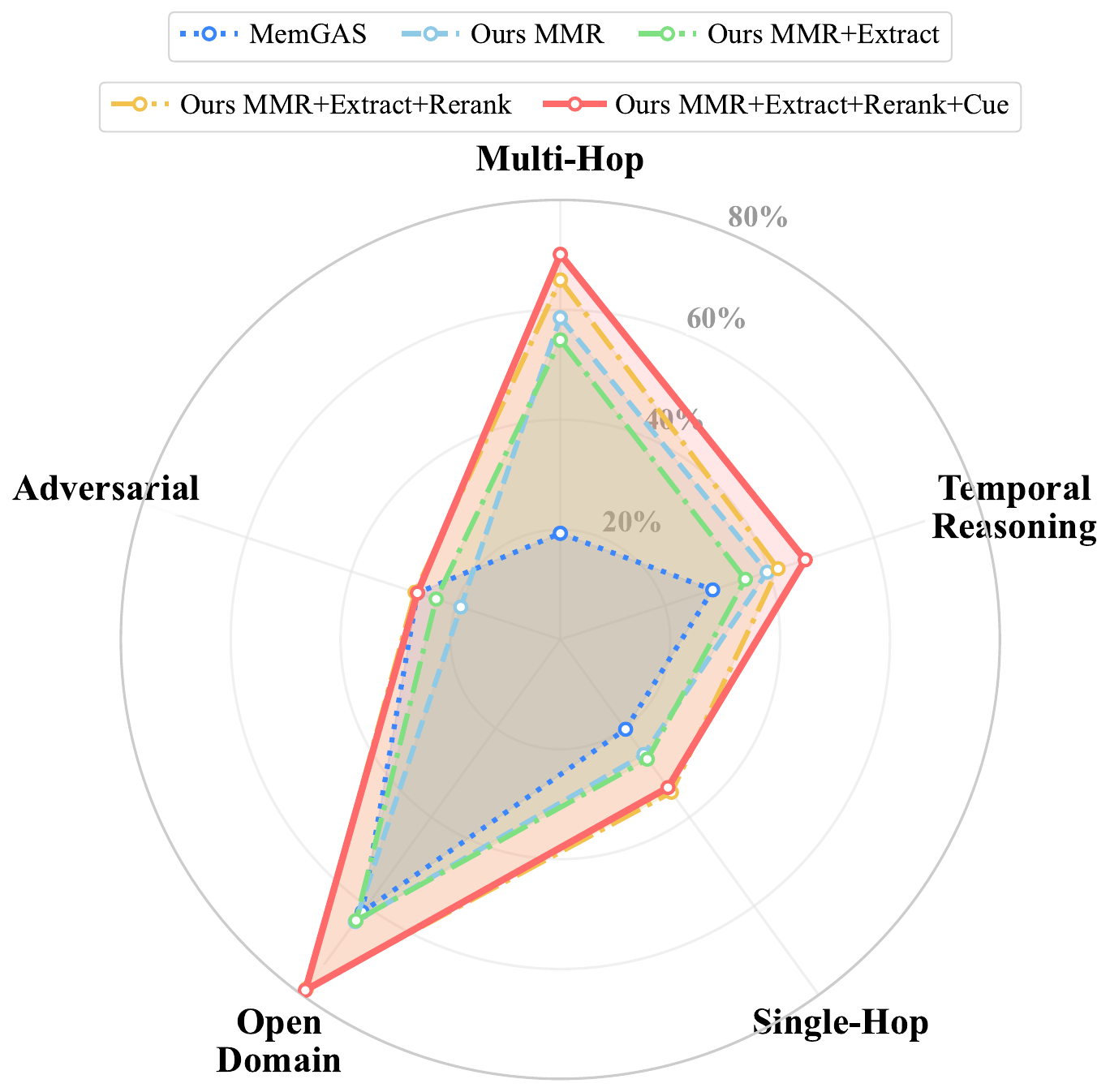}
  \caption{Performance Comparison by Query Type.}
  \label{fig:query_type}
\end{figure}

%

To comprehensively evaluate our method’s generation performance across different query types,
we conduct a detailed query-type analysis on the LoCoMo dataset. As shown in Fig.~\ref{fig:query_type}, \model consistently outperforms baselines across all query types.
Our MMR variant nearly surpasses MemGAS, demonstrating that our memory retrieval mechanism effectively leverages event associations to identify relevant memory units, thereby enhancing multi-hop reasoning and knowledge integration.
Furthermore, the full MMR+REL (Extract-Rerank-Cue) configuration achieves the best results across all query types,
confirming that the trained \module successfully reranks candidate memory units and extracts critical evidence, leading to more accurate and reliable answer generation.

\section{Implementation Details}
\label{sec:impl_details}

For each query, we retrieve the Top-10 sessions via \textit{Inner-Memory Routing} and \textit{Cross-Memory Propagation}, apply \module for \textit{Inner-Memory Extraction} and \textit{Cross-Memory Reranking}, and use the reranked Top-3 sessions for answer generation.
We employ \textbf{BGE-M3}~\cite{bge_m3} as the dense retriever and \textbf{GPT-4o mini} as the backbone for all generation and extraction tasks.
All baselines use identical prompts with temperature 0, and every experiment is run three times in a zero-shot setting.
For \textit{Inner-Memory Routing}, we use a softmax temperature of $\tau=0.1$.
For \textit{Memory Graph Construction}, we link each node to its top-$K_{\text{cand}}=20$ most similar same-granularity units, fit a two-component GMM to their similarity scores, and retain only neighbors in the higher-mean component.
For \textit{Cross-Memory Propagation}, we sparsify the seed relevance vector to its top-$K_0=15$ entries and set the PPR restart probability to $d=0.1$.

\begin{table*}[ht]
\centering
\caption{Retrieval performance with different encoders on the LoCoMo dataset.}
\label{tab:encoder_full}
\resizebox{\textwidth}{!}{
\begin{tabular}{lcccccc}
\toprule
\textbf{Model} & \textbf{Recall@3} & \textbf{NDCG@3} & \textbf{Recall@5} & \textbf{NDCG@5} & \textbf{Recall@10} & \textbf{NDCG@10} \\
\midrule

\rowcolor{blue!5}
\multicolumn{7}{c}{\textbf{Base Retriever: MiniLM}} \\
MiniLM & 42.55 & 44.19 & 52.37 & 49.01 & 67.98 & 54.59 \\
RecurSum & 44.76 & 46.82 & 54.73 & 51.64 & 72.16 & 57.52 \\
SeCom & 45.77 & 47.25 & 54.93 & 51.70 & 71.15 & 57.37 \\
MemGAS & 47.73 & 49.11 & 56.60 & 53.46 & 71.30 & 58.59 \\
\model (Ours w/o Locator) & \textbf{53.19} & 42.16 & \textbf{64.76} & 46.90 & \textbf{80.79} & 52.61 \\

\midrule

\rowcolor{blue!5}
\multicolumn{7}{c}{\textbf{Base Retriever: MPNet}} \\
MPNet & 45.92 & 47.71 & 53.98 & 51.79 & 68.58 & 56.88 \\
RecurSum & 49.50 & 51.15 & 59.47 & 56.16 & 76.64 & 61.99 \\
SeCom & 47.53 & 49.03 & 57.05 & 53.57 & 70.90 & 58.57 \\
MemGAS & 52.77 & 54.63 & 62.79 & 59.56 & 80.51 & 65.48 \\
\model (Ours w/o Locator) & \textbf{56.77} & 43.82 & \textbf{67.80} & 48.35 & \textbf{82.96} & 53.87 \\

\midrule

\rowcolor{blue!5}
\multicolumn{7}{c}{\textbf{Base Retriever: Contriever}} \\
Contriever & 49.90 & 52.15 & 58.26 & 56.29 & 71.80 & 60.92 \\
RecurSum & 47.23 & 48.99 & 59.01 & 54.58 & 74.97 & 60.07 \\
SeCom & 55.24 & 57.90 & 64.80 & 62.36 & 78.30 & 66.97 \\
MemGAS & 57.30 & 58.76 & 67.32 & 63.62 & 81.82 & 68.42 \\
\model (Ours w/o Locator) & \textbf{64.56} & 51.21 & \textbf{74.42} & 55.32 & \textbf{88.57} & 60.54 \\

\midrule

\rowcolor{blue!5}
\multicolumn{7}{c}{\textbf{Base Retriever: BGE-M3}} \\
BGE-M3 & 51.86 & 47.45 & 60.66 & 51.08 & 75.54 & 56.18 \\
RecurSum & 58.96 & 52.94 & 69.71 & 57.38 & 84.35 & 62.44 \\
SeCom & 56.11 & 49.75 & 66.59 & 54.11 & 82.27 & 59.50 \\
MemGAS & 60.62 & 52.60 & 71.46 & 57.14 & 86.07 & 62.22 \\
\model (Ours w/o Locator) & \textbf{66.84} & 54.58 & \textbf{77.10} & 58.75 & \textbf{90.09} & 63.83 \\

\bottomrule
\end{tabular}
}
\end{table*}

\paragraph{Training.}
\module is trained in two stages.

\textit{Stage~1 (SFT cold‑start):} We fine‑tune for 5 epochs with learning rate $10^{-5}$ on 4,000 multi‑hop QA samples (2–4 hops) from HotpotQA~\cite{yang2018hotpotqa}, MuSiQue~\cite{trivedi2022musique}, and 2WikiMultihopQA~\cite{ho2020constructing}.

\textit{Stage~2 (SHPO):} We adapt 6,000 conversational QA samples from Time-Dialogue~\cite{du2025memoryt1reinforcementlearningtemporal,qin-etal-2021-timedial} and HaluMem~\cite{chen2026halumemevaluatinghallucinationsmemory} to our memory-based QA.
Training uses a 3-stage GRPO (60 steps per stage, learning rate $10^{-6}$), sampling 4 rollouts per query at each step.
Queries with mixed outcomes (both correct and incorrect rollouts) yield contrastive hints, which are appended to the prompts of subsequent stages.
We set the KL coefficient $\gamma=0.01$ and the reward weights $\alpha=0.3$ and $\beta=0.6$.

\paragraph{Implementation.}
All experiments are built with Verl~\cite{zheng2025easyr1,sheng2024hybridflow} and LlamaFactory~\cite{zheng2024llamafactory} on 8$\times$A800 GPUs (PyTorch).
\textit{Long‑MT‑Bench+} is excluded from retrieval evaluation due to missing ground‑truth annotations.
For \textit{LoCoMo}, images are replaced with textual captions.
Results for \textit{RAPTOR}, \textit{A-Mem}, and \textit{HippoRAG 2} on \textit{LongMemEval-M} are unavailable due to excessive runtime.

\section{Generalization Experiments}
\label{sec:G}

\subsection{Retrieval Generalization}

As shown in Table~\ref{tab:encoder_full}, \model consistently improves retrieval performance across all base encoders, including MiniLM, MPNet, Contriever, and BGE-M3 on the LoCoMo dataset.
Importantly, the gains are stable across both lightweight and strong retrievers, indicating that our method is not dependent on any specific embedding model, but instead serves as a plug-and-play enhancement for diverse retrieval backbones.
In particular, \model improves Recall@K across all settings, demonstrating its ability to refine retrieval signals beyond the capacity of the underlying encoders.
These results demonstrate the encoder-agnostic nature and strong generalization ability of \model across heterogeneous embedding backbones.

\begin{table*}[ht]
\centering
\scriptsize
\setlength{\tabcolsep}{3pt}
\renewcommand{\arraystretch}{1.05}
\definecolor{colorBlock}{HTML}{F7F7F7}
\caption{
QA performance under the same Top-3 retrieval setting.
Best results are shown in \textbf{bold}.
}
\label{tab:qa_perf_qwen}
\resizebox{\linewidth}{!}{
\begin{tabular}{l l|cccc|cccc}
\toprule
\multirow{2}{*}{\textbf{Method}} &
\multirow{2}{*}{\textbf{Generator}} &
\multicolumn{4}{c|}{\cellcolor{blue!10}\textit{\textbf{LongMemEval-S}}} &
\multicolumn{4}{c}{\cellcolor{blue!10}\textit{\textbf{LongMemEval-M}}} \\
\cmidrule(lr){3-6}\cmidrule(lr){7-10}
& & \textbf{4o-J} & \textbf{F1} & \textbf{B-4} & \textbf{BS}
  & \textbf{4o-J} & \textbf{F1} & \textbf{B-4} & \textbf{BS} \\
\midrule

\rowcolor{white}
& \gpticon GPT-4o mini &
60.20 & {20.38} & \textbf{4.22} & \textbf{85.21} &
45.40 & 16.85 & \textbf{3.39} & \textbf{84.69}\\
\rowcolor{white}
& \qwenicon Qwen3-30B-A3B &
58.40 & \textbf{26.65} & \textbf{6.10} & \textbf{86.61} &
47.80 & \textbf{25.25} & \textbf{5.24} & \textbf{86.26} \\
\rowcolor{white}
& \qwenicon Qwen3-8B (No-Think) &
43.20 & 18.75 & 3.77 & 85.57 &
39.40 & 18.97 & 3.77 & 85.57 \\
\rowcolor{white}
\multirow{-4}{*}{\textbf{MemGAS}} & \qwenicon Qwen3-8B (Think) &
67.20 & 12.86 & \textbf{2.20} & 83.21 &
54.20 & 12.26 & 1.77 & 83.01 \\

\midrule

\rowcolor{colorBlock}
& \gpticon GPT-4o mini &
\textbf{\textcolor{red}{68.40}} & \textbf{21.52} & 3.81 & {84.33} &
\textbf{\textcolor{red}{54.20}} & \textbf{17.27} & 3.08 & 83.71\\
\rowcolor{colorBlock}
& \qwenicon Qwen3-30B-A3B &
\textbf{\textcolor{red}{71.20}} & 17.72 & 3.51 & 84.74 &
\textbf{\textcolor{red}{55.60}} & 15.88 & 2.83 & 84.27 \\
\rowcolor{colorBlock}
& \qwenicon Qwen3-8B (No-Think) &
\textbf{\textcolor{red}{58.40}} & \textbf{19.54} & \textbf{3.92} & \textbf{85.34} &
\textbf{\textcolor{red}{48.60}} & 18.20 & 3.33 & 84.99 \\
\rowcolor{colorBlock}
\multirow{-4}{*}{\textbf{\model (Ours)}} & \qwenicon Qwen3-8B (Think) &
\textbf{\textcolor{red}{73.00}} & \textbf{13.06} & 2.12 & \textbf{83.64} &
\textbf{\textcolor{red}{58.20}} & \textbf{12.38} & \textbf{1.89} & \textbf{83.38} \\

\midrule\midrule

\multirow{2}{*}{\textbf{Method}} &
\multirow{2}{*}{\textbf{Generator}} &
\multicolumn{4}{c|}{\cellcolor{blue!10}\textit{\textbf{LoCoMo}}} &
\multicolumn{4}{c}{\cellcolor{blue!10}\textit{\textbf{Long-MT-Bench+}}} \\
\cmidrule(lr){3-6}\cmidrule(lr){7-10}
& & \textbf{4o-J} & \textbf{F1} & \textbf{B-4} & \textbf{BS}
  & \textbf{4o-J} & \textbf{F1} & \textbf{B-4} & \textbf{BS} \\
\midrule

\rowcolor{white}
& \gpticon GPT-4o mini &
{41.07} & {17.66} & 3.61 & {85.13} &
{69.44} & {41.49} & \textbf{15.62} & \textbf{88.96}\\
\rowcolor{white}
& \qwenicon Qwen3-30B-A3B &
43.45 & \textbf{24.41} & \textbf{4.31} & \textbf{87.04} &
63.89 & \textbf{54.70} & 13.38 & \textbf{89.00} \\
\rowcolor{white}
& \qwenicon Qwen3-8B (No-Think) &
47.78 & 20.18 & \textbf{5.36} & \textbf{87.69} &
54.51 & \textbf{53.60} & 13.16 & 88.95 \\
\rowcolor{white}
\multirow{-4}{*}{\textbf{MemGAS}} & \qwenicon Qwen3-8B (Think) &
48.33 & \textbf{12.93} & \textbf{2.32} & 84.64 &
67.36 & 38.32 & 6.87 & 85.70 \\

\midrule

\rowcolor{colorBlock}
& \gpticon GPT-4o mini &
\textbf{\textcolor{red}{57.40}} & \textbf{19.99} & \textbf{4.26} & \textbf{85.26} &
\textbf{\textcolor{red}{77.42}} & \textbf{46.65} & 14.02 & 88.48\\
\rowcolor{colorBlock}
& \qwenicon Qwen3-30B-A3B &
\textbf{\textcolor{red}{52.82}} & 17.89 & 3.68 & 86.37 &
\textbf{\textcolor{red}{73.26}} & 48.47 & \textbf{13.94} & 88.51 \\
\rowcolor{colorBlock}
& \qwenicon Qwen3-8B (No-Think) &
\textbf{\textcolor{red}{54.92}} & \textbf{23.25} & 4.80 & 87.16 &
\textbf{\textcolor{red}{70.13}} & 50.18 & \textbf{15.07} & \textbf{89.02} \\
\rowcolor{colorBlock}
\multirow{-4}{*}{\textbf{\model (Ours)}} & \qwenicon Qwen3-8B (Think) &
\textbf{\textcolor{red}{53.82}} & 12.62 & 2.18 & \textbf{85.21} &
\textbf{\textcolor{red}{69.79}} & \textbf{40.10} & \textbf{11.30} & \textbf{87.96} \\

\bottomrule
\end{tabular}
}
\end{table*}

\subsection{Generation Generalization}

We further evaluate the generalization of \model across different generation backbones, including \gpticon GPT-4o mini, \qwenicon Qwen3-30B-A3B, and \qwenicon Qwen3-8B under both No-Think and Think settings.
As shown in Table~\ref{tab:qa_perf_qwen}, \model consistently improves QA performance over MemGAS across all datasets and backbone models, demonstrating strong cross-backbone robustness.
Notably, the improvements are observed for both strong and weak generators, indicating that \model does not rely on the reasoning capacity of the stronger generator, but instead benefits from higher-quality retrieved evidence for downstream generation.

\section{Efficiency and Latency Analysis}

\begin{table}[ht]
\centering
\caption{Efficiency and Latency comparison.}
\label{tab:latency_comparison}
\begin{threeparttable}
\resizebox{\linewidth}{!}{
\begin{tabular}{l l c c c c}
\toprule
\textbf{Dataset} & \textbf{Method} & \textbf{Tokens}
& \textbf{Filtering (s)} & \textbf{Generation (s)} & \textbf{4o-J} \\
\midrule

\multirow{3}{*}{LongMemEval-M}
& Full History & 128,000 & 0.00 & 12.88 & 12.20 \\
& MemGAS & 8,852 & 0.53 & 1.48 & 45.40 \\
& \textbf{\model} & 8,537 & 0.66 & 1.83 & \textbf{54.20} \\
\midrule

\multirow{3}{*}{LongMemEval-S}
& Full History & 103,137 & 0.00 & 9.39 & 50.60 \\
& MemGAS & 8,829 & 0.54 & 0.94 & 60.20 \\
& \textbf{\model} & 8,941 & 0.61 & 1.58 & \textbf{68.40} \\
\midrule

\multirow{3}{*}{LoCoMo}
& Full History & 20,079 & 0.00 & 4.92 & 33.43 \\
& MemGAS & 2,825 & 0.47 & 1.48 & 41.07 \\
& \textbf{\model} & 2,907 & 0.46 & 1.51 & \textbf{57.40} \\
\midrule

\multirow{3}{*}{Long-MT-Bench+}
& Full History & 19,195 & 0.00 & 4.72 & 67.44 \\
& MemGAS & 12,873 & 0.73 & 1.60 & 69.44 \\
& \textbf{\model} & 13,921 & 0.31 & 3.28 & \textbf{77.42} \\
\bottomrule
\end{tabular}
}
\begin{tablenotes}[flushleft]
\footnotesize
\setlength{\tabcolsep}{0pt}
\begin{tabular}{@{}p{\linewidth}@{}}
\item \textbf{Tokens}: Average token consumption per query.
\item \textbf{Filtering}: Average filtering latency per session.
\item \textbf{Generation}: Average answer generation latency per query.
\item \textbf{4o-J}: Measuring response quality (higher is better).
\end{tabular}
\end{tablenotes}
\end{threeparttable}
\end{table}

As shown in Table~\ref{tab:latency_comparison}, \model achieves the highest LLM-as-a-Judge (4o-J) evaluation scores across all datasets while consuming substantially fewer tokens than the Full History baseline.
Compared to the efficient MemGAS baseline, \model incurs only a marginal increase in token usage and inference latency, while consistently achieving notable gains in LLM-as-a-Judge scores.
These results suggest that our multi-granularity retrieval and evidence localization pipeline effectively compresses long-context inputs by filtering irrelevant content and retaining relevant information, thereby achieving state-of-the-art performance with low computational overhead.

\subsection{Computational Cost Analysis}
\label{sec:cost_analysis}

\begin{table}[ht]
\centering
\small
\caption{Offline indexing complexity and QA performance on LongMemEval-M ($S{\approx}500$, $N{>}250$K). ``OOT'': Out-Of-Time.}
\label{tab:complexity_compare}
\resizebox{\linewidth}{!}{
\begin{tabular}{l|c|c|c}
\toprule
\textbf{Method} & \textbf{Memory Extraction} & \textbf{Graph/Index Build} & \textbf{4o-J} \\
\midrule
Full History & -- & -- & 12.20 \\
\midrule
RAPTOR & $O(N \cdot C_{\text{LLM}})$ & $O(N^2)$ & OOT \\
A-Mem & $O(N \cdot C_{\text{LLM}})$ & $O(N^2 + N \cdot C_{\text{LLM}})$ & OOT \\
HippoRAG 2 & $O(N \cdot C_{\text{LLM}})$ & $O(N^2)$ & OOT \\
Mem0 & $O(N \cdot C_{\text{LLM}})$ & $O(N^2)$ & 32.00 \\
MemoryTree & $O(N \cdot C_{\text{LLM}})$ & $O(N \log N)$ & 35.80 \\
Memory-R1 & $O(N \cdot C_{\text{LLM}})$ & - & 29.00 \\
\midrule
SeCom & $O(S \cdot C_{\text{LLM}})$ & - & 42.80 \\
RecurSum & $O(S \cdot C_{\text{LLM}})$ & - & 23.80 \\
MemoryAgent & $O(S \cdot C_{\text{LLM}})$ & -- & 34.20 \\
MemSkill & $O(S \cdot C_{\text{LLM}})$ & - & 42.20 \\
MemGAS & $O(S \cdot C_{\text{LLM}})$ & $O(G \cdot S^2)$ & 45.40 \\
\midrule
\rowcolor{orange!5}
\textbf{\model (Ours)} & $O(S \cdot C_{\text{LLM}})$ & $O(G \cdot S^2)$ & \textbf{54.20} \\
\bottomrule
\end{tabular}
}
\end{table}

Table~\ref{tab:complexity_compare} compares the offline indexing complexity and QA performance of different memory systems on LongMemEval-M.
We decompose the offline cost into two stages: memory extraction (node construction) and graph/index construction (edge construction).
Let $S$ denote the number of dialogue sessions, $\bar{n}$ the average number of utterances per session, and $N = S \cdot \bar{n}$ the total number of utterances.
We further denote by $G$ the number of memory granularities, and $C_{\text{LLM}}$ the cost of a single LLM inference call.

\begin{table*}[th]
\centering
\scriptsize

\definecolor{colorMain}{HTML}{E6F0FF}
\definecolor{colorBlock}{HTML}{F7F7F7}

\caption{
Best results are shown in \textbf{bold}, and second-best results are \underline{underlined}.
\textbf{4o-J} refers to GPT-4o-as-Judge, \textbf{B-4} to BLEU-4, and \textbf{BS} to BERTScore.
\textbf{Tokens} denotes the average token consumption per query. All methods use the same generator \gpticon GPT-4o mini.
}

\label{tab:qa_perf_top3_all}

\resizebox{0.95\linewidth}{!}{
\begin{tabular}{l|ccccc|ccccc}
\toprule
\multirow{2}{*}{\textbf{Model}} &
\multicolumn{5}{c|}{\cellcolor{blue!10}\textit{\textbf{LongMemEval-S}}} &
\multicolumn{5}{c}{\cellcolor{blue!10}\textit{\textbf{LongMemEval-M}}} \\
\cmidrule(lr){2-6}\cmidrule(lr){7-11}
& \textbf{4o-J} & \textbf{F1} & \textbf{B-4} & \textbf{BS} & \makecell{\textbf{Tokens}}
& \textbf{4o-J} & \textbf{F1} & \textbf{B-4} & \textbf{BS} & \makecell{\textbf{Tokens}} \\
\midrule

Full History &
50.60 & 11.48 & 1.40 & 83.07 & 103,137 &
12.20 & 5.70 & 0.78 & 81.62 & 128,000 \\

RAPTOR &
32.20 & 12.08 & 1.90 & 83.50 & 6,254 &
\multicolumn{5}{c}{\cellcolor{gray!5}\textcolor{gray}{\textit{Timeout}}} \\

SeCom &
56.00 & 12.95 & 2.25 & 83.51 & 2,741 &
42.80 & 11.33 & 1.79 & 83.36 & 2,821 \\

A-Mem &
55.60 & 13.73 & 2.11 & 83.88 & 9,018 &
\multicolumn{5}{c}{\cellcolor{gray!5}\textcolor{gray}{\textit{Timeout}}} \\

HippoRAG 2 &
57.60 & 14.73 & 2.15 & 83.86 & 8,530 &
\multicolumn{5}{c}{\cellcolor{gray!5}\textcolor{gray}{\textit{Timeout}}} \\

Mem0 &
42.00 & 17.72 & 3.62 & 83.25 & 6,787 &
32.00 & 14.38 & 2.57 & 82.34 & 7,452 \\

MemGAS &
60.20 & 20.38 & \underline{4.22} & 85.21 & 8,829 &
45.40 & \underline{16.85} & \textbf{3.39} & \underline{84.69} & 8,852 \\

MemoryAgent &
42.80 & 11.71 & 3.91 & 81.48 & 11,200 &
34.20 & 9.58 & 2.90 & 81.33 & 11,800 \\

Memory-R1 &
45.60 & \underline{21.49} & \textbf{8.66} & \textbf{87.06} & 13,500 &
29.00 & 5.95 & 1.06 & 80.55 & 14,200 \\

RecurSum &
35.40 & 12.29 & 2.09 & 83.60 & 8,853 &
23.80 & 10.04 & 1.70 & 83.12 & 8,927 \\

MemoryTree &
41.40 & 12.02 & 2.45 & 82.90 & 7,600 &
35.80 & 11.57 & 2.23 & 83.08 & 8,200 \\

MemSkill &
59.14 & 15.61 & 3.66 & 84.35 & 6,900 &
42.20 & 14.88 & 3.42 & 84.07 & 7,300 \\

\rowcolor{orange!5}
\tabmark \textbf{\model (Ours, 4B)} &
\underline{65.80} & 16.80 & 3.78 & \underline{85.40} & 8,645 &
\underline{48.20} & 13.41 & 2.90 & \underline{84.71} & 8,326 \\

\rowcolor{orange!5}
\tabmark \textbf{\model (Ours, 8B)} &
\textbf{68.40} & \textbf{21.52} & 3.81 & 84.33 & 8,941 &
\textbf{54.20} & \textbf{17.27} & \underline{3.08} & 83.71 & 8,537 \\

\midrule\midrule

\multirow{2}{*}{\textbf{Model}} &
\multicolumn{5}{c|}{\cellcolor{blue!10}\textit{\textbf{LoCoMo}}} &
\multicolumn{5}{c}{\cellcolor{blue!10}\textit{\textbf{Long-MT-Bench+}}} \\
\cmidrule(lr){2-6}\cmidrule(lr){7-11}
& \textbf{4o-J} & \textbf{F1} & \textbf{B-4} & \textbf{BS} & \makecell{\textbf{Tokens}}
& \textbf{4o-J} & \textbf{F1} & \textbf{B-4} & \textbf{BS} & \makecell{\textbf{Tokens}} \\
\midrule

Full History &
33.43 & 12.23 & 1.84 & 84.07 & 20,079 &
67.44 & 36.07 & 11.32 & 87.81 & 19,195 \\

RAPTOR &
31.72 & 14.55 & 2.88 & 84.48 & 1,931 &
59.72 & 37.69 & 13.47 & 88.38 & 10,631 \\

SeCom &
44.21 & 13.79 & 2.30 & 84.04 & 1,021 &
64.58 & 36.68 & 12.01 & 87.88 & 4,714 \\

A-Mem &
40.81 & 14.72 & 2.83 & 84.72 & 3,042 &
65.73 & 36.82 & 11.36 & 87.92 & 13,735 \\

HippoRAG 2 &
45.62 & 16.66 & 2.91 & 84.88 & 2,991 &
63.54 & 35.64 & 11.05 & 87.70 & 13,583 \\

Mem0 &
36.39 & 8.16 & 2.06 & 83.70 & 2,450 &
38.33 & 25.42 & 6.06 & 86.86 & 8,100 \\

MemGAS &
41.07 & 17.66 & 3.61 & 85.13 & 2,825 &
69.44 & \underline{41.49} & \textbf{15.62} & \textbf{88.96} & 12,873 \\

MemoryAgent &
38.72 & 14.80 & \textbf{4.76} & 81.75 & 6,800 &
39.03 & 11.69 & 2.06 & 82.48 & 16,200 \\

Memory-R1 &
38.07 & 5.52 & 1.01 & 81.32 & 7,900 &
55.90 & 8.46 & 1.41 & 81.37 & 18,400 \\

RecurSum &
22.56 & 9.14 & 0.99 & 83.45 & 3,074 &
24.65 & 26.58 & 6.91 & 86.11 & 13,527 \\

MemoryTree &
35.21 & 8.95 & 1.62 & 83.49 & 5,300 &
51.39 & 30.13 & 8.55 & 86.99 & 11,500 \\

MemSkill &
47.21 & 17.37 & 3.37 & \underline{85.43} & 4,900 &
62.33 & 32.01 & 9.89 & 87.44 & 10,700 \\

\rowcolor{orange!5}
\tabmark \textbf{\model (Ours, 4B)} &
\underline{57.20} & \underline{18.13} & 4.13 & \textbf{86.88} & 2,744 &
\underline{73.96} & 37.98 & 13.70 & \underline{88.49} & 12,463 \\

\rowcolor{orange!5}
\tabmark \textbf{\model (Ours, 8B)} &
\textbf{57.40} & \textbf{19.99} & \underline{4.26} & 85.26 & 2,907 &
\textbf{77.42} & \textbf{46.65} & \underline{14.02} & 88.48 & 13,921 \\

\bottomrule
\end{tabular}
}
\end{table*}

\paragraph{Analysis.}
\textbf{(1) Session-level extraction substantially reduces offline indexing cost.}
Methods such as RAPTOR, A-Mem, HippoRAG~2, Mem0, MemoryTree, and Memory-R1 perform memory extraction at the utterance level, resulting in $O(N \cdot C_{\text{LLM}})$ complexity.
On LongMemEval-M, where $N{>}250$K, this requires a prohibitively large number of LLM calls, causing several methods to exceed the runtime budget.
In contrast, \model performs joint multi-granularity extraction with a single LLM call per session, reducing the extraction cost to $O(S \cdot C_{\text{LLM}})$.
%
\textbf{(2) Graph construction introduces moderate overhead but yields substantial gains.}
Both MemGAS and \model construct cross-session relational graphs with complexity $O(G \cdot S^2)$ based on embedding similarity.
Unlike methods that require LLM inference during graph construction, our pipeline operates entirely on pre-computed embeddings.
Although graph construction incurs additional offline computation compared with flat retrieval systems such as SeCom and RecurSum, the resulting relational structure enables effective cross-memory propagation and multi-hop evidence association across sessions.
\textbf{(3) Structured graph indexing substantially improves QA quality.}
Flat retrieval methods without explicit relational modeling achieve limited performance despite lower indexing cost.
In contrast, \model achieves the best overall QA performance while maintaining computational complexity comparable to existing graph-based methods.
These results demonstrate that incorporating structured cross-memory graph construction enables more effective evidence association and long-context reasoning, yielding substantial gains in long-term conversation QA quality with only moderate offline overhead.

\section{LLM-as-Judge and Human Evaluation}
\label{app:judge_human_eval}

We use GPT-4o as the primary judge, following prior work including MemSkill, MemGAS, and Memory-R1. LLM-based evaluation is particularly important for long-term memory QA, where semantically equivalent answers can differ substantially in surface form. For example, in LoCoMo, the question ``How many doctor's appointments did I go to in March?'' has the ground-truth answer ``2'', while our model produces ``Two doctors, the first is dentist Tom, the second is dermatologist Mike.'' Although semantically correct, the token-level F1 score is only 7.7. This discrepancy is further supported by the BERTScore results in Appendix~\ref{sec:G} (Table~\ref{tab:qa_perf_top3_all}), where high semantic similarity can coexist with low lexical overlap.

\begin{table}[ht]
\centering
\caption{LLM-as-judge evaluation on LongMemEval-S using different judges. Scores are percentages.}
\label{tab:multi_judge}
\begin{adjustbox}{width=\linewidth,center}
\resizebox{\columnwidth}{!}{
\begin{tabular}{@{} l c c c c @{}}
\toprule
\textbf{Method}
& \textbf{GPT-4o}
& \textbf{GPT-5}
& \textbf{Claude-Sonnet 4.6}
& \textbf{DeepSeek-V4-Pro} \\
\midrule
\textbf{MemLoc (Ours)}
& \textbf{68.40} & \textbf{69.60} & \textbf{70.00} & \textbf{66.80} \\
MemGAS
& 60.20 & 60.80 & 60.00 & 58.20 \\
MemSkill
& 59.14 & 55.80 & 56.40 & 54.60 \\
Memory-R1
& 45.60 & 41.20 & 41.80 & 41.20 \\
MemoryAgent
& 42.80 & 41.80 & 43.40 & 39.80 \\
Mem0
& 42.00 & 39.80 & 40.40 & 38.40 \\
MemoryTree
& 41.40 & 42.60 & 43.20 & 40.40 \\
\bottomrule
\end{tabular}
}
\end{adjustbox}
\end{table}

To examine potential judge bias, we additionally evaluate LongMemEval-S responses using GPT-5, Claude-Sonnet 4.6, and DeepSeek-V4-Pro. As shown in Table~\ref{tab:multi_judge}, MemLoc consistently ranks first across all four judges, with an 8--10 point margin over the strongest baseline. The variation across judges is also small, suggesting that the observed improvements are robust to the choice of evaluator.

We further conduct human evaluation on all 500 LongMemEval-S responses. Two annotators independently assess each response, with a third annotator resolving disagreements. The resulting Cohen's $\kappa=0.92$ indicates almost perfect inter-annotator agreement. Human evaluation is consistent with the LLM-based results and confirms that MemLoc significantly outperforms the compared baselines.

\section{Prompt Templates}
We provide the prompt templates used at different stages of our framework.
In the templates, \textcolor{blue}{blue text} denotes sample-specific inputs, while black text represents fixed instructions shared across all instances.
Since our framework is model-agnostic, these prompts can be freely integrated with different LLMs and readily benefit from future advances in foundation models.

\definecolor{confblue}{RGB}{31, 78, 121}
\definecolor{confbluebg}{RGB}{245, 248, 252}

\begin{tcolorbox}[
    float*=ht,
    title=Prompt Template for Summarization and Keyword Extraction,
    width=\textwidth,
    colframe=confblue,
    colback=confbluebg,
    colbacktitle=confblue,
    coltitle=white,
    fonttitle=\bfseries,
    boxrule=0.8pt,
    arc=1.5pt,
    breakable
]
    \small
    \textbf{Instruction:} You are an intelligent and insightful individual. Your task is to analyze the conversation between a user and an AI assistant and extract two key elements: a concise summary of the conversation and the most relevant keywords. \\

    \textbf{Input:} \textcolor{blue}{\{Conversation between User and AI Assistant\}} \\

    \textbf{Tasks:} \\
    1. \textbf{Summary}: Provide a concise paragraph that summarizes the main topics and key information of the conversation. \\
    2. \textbf{Keywords}: Extract the most relevant keywords from the conversation content. \\

    \textbf{Output Format:} \\
    Return a JSON object that must strictly contain the following structure:
    \begin{verbatim}
{
    "memory":
        {
        "summary": "<A concise summary of the conversation>",
        "keywords": "<Keyword 1>; <Keyword 2>; <Keyword 3>; ..."
        }
}
    \end{verbatim}

    Only provide the JSON object without any additional text. \\

    \textbf{Answer:}
\end{tcolorbox}

\begin{tcolorbox}[
    float*=ht,
    title=Prompt Template for Event Extraction and Timeline Construction,
    width=\textwidth,
    colframe=confblue,
    colback=confbluebg,
    colbacktitle=confblue,
    coltitle=white,
    fonttitle=\bfseries,
    boxrule=0.8pt,
    arc=1.5pt,
    breakable
]
    \small
    \textbf{Instruction:} You are an intelligent system designed to extract structured temporal information from conversations. Your task is to identify key events and construct a standardized timeline based on the given dialogue. \\

    \textbf{Input:} \\
    \textbf{Conversation Date:} \textcolor{blue}{\{Conversation Date\}} \\
    \textbf{Conversation Content:} \textcolor{blue}{\{Conversation between User and AI Assistant\}} \\

    \textbf{Tasks:} \\
    1. \textbf{Key Event Extraction}: Identify the key events mentioned in the conversation. \\
    2. \textbf{Timeline Construction}: Organize the extracted events in chronological order. \\
    3. \textbf{Role Awareness}: Summarize the information provided by the \textbf{assistant} and the \textbf{user} separately. \\

    \textbf{Requirements:} \\
    - Output only concise event summaries. \\
    - Do NOT quote or restate the original dialogue. \\
    - Ensure the results are clear, structured, and information-dense. \\

    \textbf{Output Format:} \\
    \begin{verbatim}
| Date | Key Event | Description |
|------|-----------|-------------|
| YYYY/MM/DD | xxx | xxx |
| YYYY/MM/DD | xxx | xxx |
    \end{verbatim}

    \textbf{Answer:}
\end{tcolorbox}

\begin{tcolorbox}[
    float*=ht,
    title=Prompt Template for Evidence Filtering and Sentence Selection,
    width=\textwidth,
    colframe=confblue,
    colback=confbluebg,
    colbacktitle=confblue,
    coltitle=white,
    fonttitle=\bfseries,
    boxrule=0.8pt,
    arc=1.5pt,
    breakable
]
    \small
    \textbf{Instruction:} You are an expert in information retrieval and evidence filtering. Your task is to identify the sentences that are most relevant to answering a given question based on retrieved documents. \\

    \textbf{Input:} \\
    \textbf{Question:} \textcolor{blue}{\{Question\}} \\
    \textbf{Retrieved Documents:} \textcolor{blue}{\{List of sentences with IDs\}} \\

    \textbf{Tasks:} \\
    1. Analyze the relevance of each sentence to the question. \\
    2. Select the most relevant sentences. \\
    3. Provide a final answer based on the selected evidence. \\

    \textbf{Requirements:} \\
    - Perform step-by-step reasoning. \\
    - Only select the most relevant sentences. \\
    - Ensure the final answer is accurate and grounded in the selected evidence. \\

    \textbf{Output Format:} \\
    \begin{verbatim}
<reason>...step-by-step reasoning...</reason>
<id>Most relevant sentence IDs</id>
<answer>answer</answer>
    \end{verbatim}

    \textbf{Answer:}
\end{tcolorbox}

\begin{tcolorbox}[
    float*=ht,
    title=Prompt Template for Multi-Granular Reasoning and Answer Generation,
    width=\textwidth,
    colframe=confblue,
    colback=confbluebg,
    colbacktitle=confblue,
    coltitle=white,
    fonttitle=\bfseries,
    boxrule=0.8pt,
    arc=1.5pt,
    breakable
]
    \small
    \textbf{Instruction:} You are an intelligent conversational assistant. Your task is to carefully analyze the provided conversation history and multi-granular contextual information, and generate a concise, accurate, coherent, and helpful answer to the given question. \\

    \textbf{Input:} \\
    \textbf{Conversation History and Context:} \textcolor{blue}{\{Retrieved texts with global IDs\}} \\
    \textbf{Reference Information (optional):} \textcolor{blue}{\{Global ID mapping\}} \\
    \textbf{Question:} \textcolor{blue}{\{Question\}} \\

    \textbf{Tasks:} \\
    1. Understand the conversation history and contextual signals. \\
    2. Integrate multi-granular information (e.g., session, summary, events, time). \\
    3. Generate a coherent and accurate answer. \\

    \textbf{Requirements:} \\
    - The answer must be grounded primarily in the provided conversation history. \\
    - Do NOT introduce unsupported or external information. \\
    - Keep the answer concise and focused. \\
    - Ensure logical coherence and readability. \\
    - If the question involves time, provide explicit timestamps when possible. \\
    - Prefer reusing original wording from the conversation when appropriate. \\
    - Use reference information only as supporting clues, not as the sole basis. \\
    - Answer must be in English. \\

    \textbf{Output Format:} \\
    Provide a direct and concise answer to the question. \\

    \textbf{Answer:}
\end{tcolorbox}

\begin{tcolorbox}[
    float*=ht,
    title=Prompt Template for Answer Verification,
    width=\textwidth,
    colframe=confblue,
    colback=confbluebg,
    colbacktitle=confblue,
    coltitle=white,
    fonttitle=\bfseries,
    boxrule=0.8pt,
    arc=1.5pt,
    breakable
]
    \small
    \textbf{Instruction:} Determine whether the model response correctly answers the question based on the reference answer. \\

    \textbf{Input:} \\
    \textbf{Question:} \textcolor{blue}{\{Question\}} \\
    \textbf{Reference Answer:} \textcolor{blue}{\{Answer\}} \\
    \textbf{Model Response:} \textcolor{blue}{\{Response\}} \\

    \textbf{Requirement:} \\
    Output \texttt{[[yes]]} if the response is correct or equivalent to the reference answer; otherwise output \texttt{[[no]]}. \\

    \textbf{Output Format:} \\
\begin{verbatim}
[[yes]] or [[no]]
\end{verbatim}

    \textbf{Answer:}
\end{tcolorbox}

\begin{tcolorbox}[
    float*=ht,
    title=Prompt Template for SFT \& RL Training,
    width=\textwidth, 
    colframe=confblue,
    colback=confbluebg,
    colbacktitle=confblue,
    coltitle=white,
    fonttitle=\bfseries,
    boxrule=0.8pt,
    arc=1.5pt,
    breakable
]
    \small
    \textbf{Instruction:} You are an expert in information retrieval and evidence filtering.  Given a question and a list of retrieved documents, select the documents that are most relevant to answering the question.  Output the corresponding document IDs, answer and explain your reasoning. \\

    \textbf{Input:} \\
    \textbf{Question:} \textcolor{blue}{\{Question\}} \\
    \textbf{Retrieved Documents:} \textcolor{blue}{\{Documents\}} \\
    
    \textbf{Output Format:}  \\
    <reason>...step-by-step reasoning...</reason> \\
    <id>Most relevant document IDs</id> \\
    <answer>answer</answer> \\
\end{tcolorbox}

\begin{tcolorbox}[
    float*=ht,
    title=Prompt Template for Hint Extraction,
    width=\textwidth,
    colframe=confblue,
    colback=confbluebg,
    colbacktitle=confblue,
    coltitle=white,
    fonttitle=\bfseries,
    boxrule=0.8pt,
    arc=1.5pt,
    breakable
]
    \small
    \textbf{Instruction:} You are an intelligent and analytical teacher model. Your task is to analyze the divergence between correct reasoning trajectories and wrong reasoning trajectories, then extract a concise and domain-general corrective hint without leaking the ground-truth answer. \\

    \textbf{Input:}\\
    \textbf{Question:} \textcolor{blue}{\{Question\}} \\
    \textbf{Correct Reasoning Trace:} \textcolor{blue}{\{Correct Reasoning Trace\}} \\
    \textbf{Wrong Reasoning Trace:} \textcolor{blue}{\{Wrong Reasoning Trace\}} \\
    \textbf{Answer:} \textcolor{blue}{\{Answer\}} \\

    \textbf{Tasks:} \\
    1. \textbf{Identify Divergence}: Determine which reasoning step caused the student model to deviate from the correct reasoning path. \\
    2. \textbf{Generate Hint}: Produce a concise, domain-general strategy that can guide the student model back to the correct reasoning direction. The hint should focus on reasoning principles rather than specific answers. \\
    3. \textbf{Leakage Verification}: Ensure that the generated hint does not reveal or directly imply the ground-truth answer. \\

    \textbf{Example Hints:}
    \begin{itemize}
        \item When multiple timestamps conflict, prioritize the most recent entry as the user's current state.
        \item Verify whether the event occurred or was merely planned. Distinguish intent from completion.
        \item Disregard assistant-generated summaries when they contradict explicit user statements.
        \item Cross-check person attributes (age, location) across sessions; use the latest update.
    \end{itemize}

    \textbf{Output Format:} \\
    Return a JSON object that must strictly contain the following structure:
\begin{verbatim}
{
    "hint":
    {
        "divergence_step": "<Reasoning step causing divergence>",
        "strategy_hint": "<Domain-general corrective hint>"
    }
}
\end{verbatim}

    Only provide the JSON object without any additional text. \\

    \textbf{Answer:}
\end{tcolorbox}

\end{document}